%% file: neurips_2024.tex
\documentclass{article}

\usepackage[preprint]{neurips_2024}

\usepackage[utf8]{inputenc} 
\usepackage[T1]{fontenc}    
\usepackage{hyperref}       
\usepackage{url}            
\usepackage{booktabs}       
\usepackage{amsfonts}       
\usepackage{nicefrac}       
\usepackage{microtype}      
\usepackage[table]{xcolor}
\usepackage{minted}
\usepackage[breakable]{tcolorbox}
\tcbuselibrary{listings, skins}
\usepackage{wrapfig}
\usepackage{subcaption}
\usepackage{caption}
\usepackage{balance}
\usepackage{url}
\usepackage{graphicx}
\usepackage{amsmath}
\usepackage{amsthm}
\usepackage{booktabs}
\usepackage{algorithm}
\usepackage{algorithmic}
\usepackage{color}
\usepackage{xcolor}
\usepackage{xspace}
\usepackage[misc]{ifsym}
\usepackage{amsmath}
\usepackage{amssymb}
\usepackage{empheq}    
\newcommand{\model}{PACE\xspace}
\newcommand{\fullmodel}{\textbf{P}refix-Protected and Difficulty-\textbf{A}ware \textbf{C}ompression for \textbf{E}fficient Reasoning\xspace}

\usepackage[ruled, vlined, nofillcomment, linesnumbered, algo2e]{algorithm2e}
\usepackage{booktabs}
\usepackage{natbib} 
\usepackage{multirow}
\usepackage{balance}
\usepackage{enumitem}
\usepackage{stfloats}
\usepackage{diagbox}
\usepackage{fancyhdr}
\definecolor{result_color}{RGB}{250,250,210}
\usepackage{bm}
\usepackage[table]{xcolor}

\newcommand{\eg}{\emph{e.g.,}\xspace}

\newcommand{\ignore}[1]{}

\title{%
  \texorpdfstring{%
    \raisebox{-0.2\height}{\includegraphics[height=1.2em]{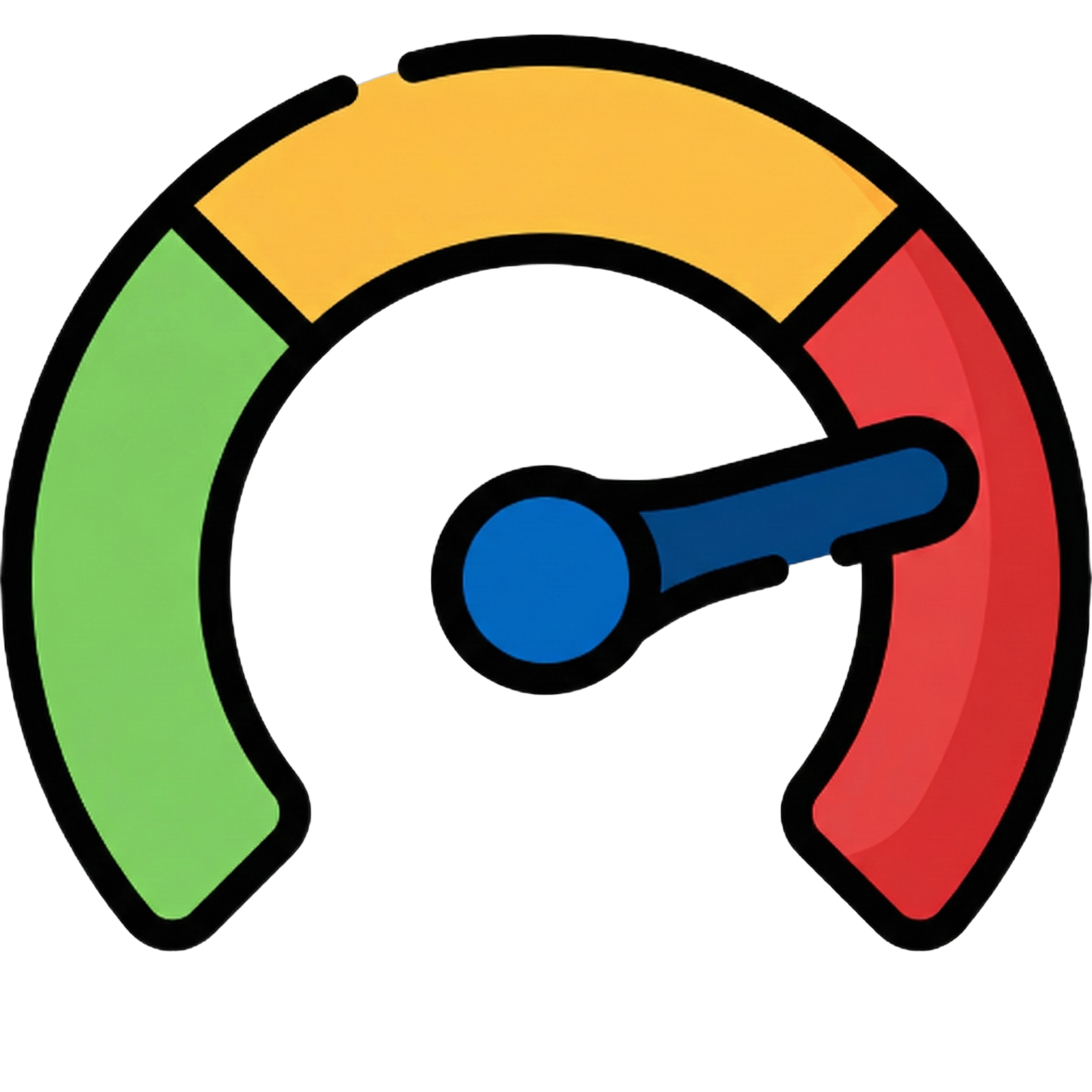}}%
    \hspace{0.4em}%
  }{}%
  \textsc{PACE}: Prefix-Protected and Difficulty-Aware Compression for Efficient Reasoning
}

\author{%
  \parbox{\linewidth}{\centering
    Ruixiang Feng\textsuperscript{1,2}\thanks{Equal contribution.}\;
    Yuntao Wen\textsuperscript{1,2}\footnotemark[1],\;
    Silin Zhou\textsuperscript{1}\footnotemark[1],\;
    Ke Shi\textsuperscript{1},\;
    Yifan Wang\textsuperscript{1},\;
    Ran Le\textsuperscript{2},\;
    Zhenwei An\textsuperscript{2},\\
    Zongchao Chen\textsuperscript{2},\;
    Chen Yang\textsuperscript{2},\;
    Guangyue Peng\textsuperscript{3},\;
    Yiming Jia\textsuperscript{2},\;
    Dongsheng Wang\textsuperscript{1},\\
    Tao Zhang\textsuperscript{2},\;
    Lisi Chen\textsuperscript{1},\;
    Yang Song\textsuperscript{2}\thanks{Corresponding author.},\;
    Shen Gao\textsuperscript{1}\footnotemark[2],\;
    Shuo Shang\textsuperscript{1}\footnotemark[2]
    \\[0.6em]
    \normalfont
    \textsuperscript{1}University of Electronic Science and Technology of China\\
    \textsuperscript{2}Nanbeige Lab, BOSS Zhipin\\[0.2em]
    \textsuperscript{3}Peking University\\[0.4em]
    \texttt{%
      \{fengrx02,yuntaowenx,jedi.shang\}@gmail.com;\ %
      \{shengao,chenlisi\}@uestc.edu.cn;\ %
      \{songyang,leran,anzhenwei\}@kanzhun.com%
    }
  }%
}

\begin{document}

\noindent\includegraphics[height=0.8cm]{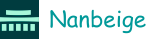}

\maketitle

\begin{abstract}
Language Reasoning Models (LRMs) achieve strong performance by scaling test-time computation but often suffer from ``overthinking'', producing excessively long reasoning traces that increase latency and memory usage. 
Existing LRMs typically enforce conciseness with uniform length penalties, which over-compress crucial early deduction steps at the sequence level and indiscriminately penalize all queries at the group level. To solve these limitations, we propose \textbf{\model}, a dual-level framework for prefix-protected and difficulty-aware compression under hierarchical supervision. At the sequence level, prefix-protected optimization employs decaying mixed rollouts to maintain valid reasoning paths while promoting conciseness. At the group level, difficulty-aware penalty dynamically scales length constraints based on query complexity, maintaining exploration for harder questions while curbing redundancy on easier ones. Extensive experiments on DeepSeek-R1-Distill-Qwen (1.5B/7B) demonstrate that \model achieves a substantial reduction in token usage (up to \textbf{55.7\%}) while simultaneously improving accuracy (up to \textbf{4.1\%}) on math benchmarks, with generalization ability to code, science, and general domains\footnote{Code and model are available at \url{https://github.com/JackWenx/PACE}.}.
\end{abstract}

\input{sec_introduction}

\input{sec_relatedwork}

\input{sec_method}

\input{sec_experiment}

\input{sec_conclusion}

\bibliographystyle{unsrt} 
\bibliography{references}

\newpage
\appendix
\input{sec_appendix}

\end{document}

%% file: sec_introduction.tex
\section{Introduction}
Recently, Large Language Models (LLMs) have evolved toward Language Reasoning Models (LRMs), which improve performance by scaling test-time computation~\cite{muennighoff2025s1, khatri2025art}.
LRMs can perform extended multi-step reasoning and thus demonstrate remarkable performance improvements on logical reasoning tasks, such as mathematics~\cite{chen2025seed, shao2024deepseekmath}, tool using~\cite{zhang2025nemotron,huang2025ttpa,hao2026failure,feng2025cosineverifier}, and software engineering~\cite{pan2024training, yang2025swe}. 

However, extended generation sequences impose a severe computational overhead. 
As a result, inference is increasingly memory-bound, with the Key-Value (KV) cache scaling rapidly with sequence length, leading to higher latency and increased GPU memory consumption~\cite{kang2024gear}. 
Moreover, long generation sequences often suffer from an \emph{overthinking} issue~\cite{sui2025stopoverthinking, chen2025donotthinkthatmuch,fan2025cothink}, where LRMs produce excessively long and unnecessary reasoning steps even for simple problems. 
For example, a standard LLM solves a basic equation ``$8x + 9 = 23, x=?$'' in around 50 tokens, whereas an LRM (e.g., DeepSeek-R1~\cite{guo2025deepseek}) may use over 1,000 tokens for the same answer, incurring substantial redundancy and wasting computation.

\begin{figure}[t]
        \centering
	    \includegraphics[width=0.5\linewidth]{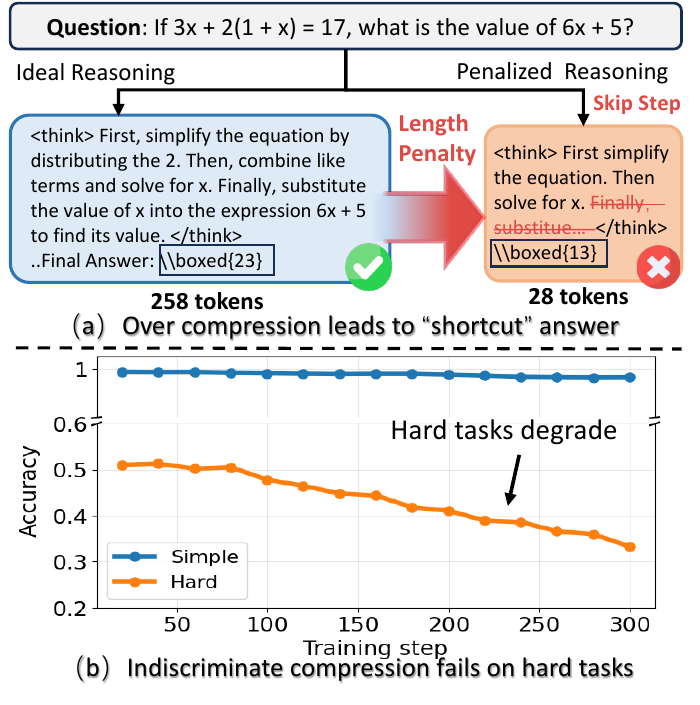}
        \caption{Two limitations of uniform length penalties. 
        (a) \textbf{Over compression}: Using a uniform length penalty will skip steps and lead to a ``shortcut'' answer. 
        (b) \textbf{Indiscriminative compression}: We report avg pass@1 (32 samples) and split queries into \emph{Simple} ($>0.75$) and \emph{Hard} ($\le 0.75$) on \textsc{MATH500}; \emph{Hard} accuracy drops sharply during training.}
 \label{fig:intro}
 \vspace{-20pt}
\end{figure}

Recent works compress long reasoning steps into shorter ones by introducing a simple length penalty during Reinforcement Learning (RL)~\cite{arora2025trainingreasoningeffectively,luo2025o1}.
However, we find this simple but rigid constraint introduces limitations at two distinct levels.
First, at the sequence level, the uniform penalty leads to \textbf{over compression} problem, as applying the same constraint to \emph{every token} prevents LRMs from distinguishing between redundant and essential steps, particularly in the critical early stages~\cite{kimiteam2025kimik15, yuan2025efficient}. 
Under this rigid penalty, LRMs may skip critical reasoning steps or simply avoid reasoning, leading to ``shortcut'' answers, reduced accuracy, and unstable training. 
As shown in Figure~\ref{fig:intro}(a), a uniform length penalty compresses the reasoning from 258 to 28 tokens, greatly improving efficiency but skipping essential steps and leading to incorrect answers.
Second, at the group level, a uniform penalty causes \textbf{indiscriminative compression} problem, as they apply the same constraint to \emph{every query}. 
This ignores the fact that complex tasks inherently require more reasoning steps, disproportionately shortening reasoning for difficult tasks and limiting LRMs' ability to perform the extended exploration and reflection necessary for correct solutions. 
Figure~\ref{fig:intro}(b) shows accuracy trends under a uniform length penalty across different difficulty levels. 
As training steps progress, accuracy remains high on simple tasks but drops sharply on hard ones.

To address these limitations, we propose a dual-level framework \textbf{\model}, which incorporates \fullmodel under hierarchical supervision. To tackle the over compression issue at the sequence level, we introduce \textit{prefix-protected optimization}. 
Specifically, we utilize a prefix rollout strategy that samples initial reasoning steps from a frozen reference policy to explicitly anchor valid solution paths. 
By gradually removing this safeguard through a linear decay schedule, we prevent the reasoning collapse often caused by premature length penalization, ensuring the model preserves logical coherence while progressively learning conciseness. For indiscriminative compression issue at the group level, we propose a \textit{difficulty-aware penalty}. 
Unlike uniform length constraints, this method first performs difficulty quantification by computing the empirical pass rate across group rollouts. 
Subsequently, it applies adaptive penalty modulation to penalty: constraints are effectively relaxed for difficult (low-pass-rate) tasks to accommodate necessary exploration, while strictly enforced for simple tasks to maximize efficiency.

Extensive experiments show that \model reduces token usage by an average of 55\% across multiple benchmarks and also maintains the reasoning performance without any degradation. Out-of-domain tasks further confirm \model generalization with shorter generations and better performance.

\noindent Our contributions are summarized as follows:

\noindent $\bullet$ We propose a dual-level framework \model to tackle the over compression and indiscriminative compression issue of efficient reasoning methods.

\noindent $\bullet$ We introduce prefix-protected optimization, using prefix rollout to anchor reasoning paths and prevent sequence-level over compression.

\noindent $\bullet$ We propose difficulty-aware penalty, scaling length penalties with task difficulty to prevent group-level indiscriminate compression.

\noindent $\bullet$ Experiments show that \model achieves the best efficiency without accuracy loss compared to other methods. It also generalizes well to OOD tasks, shows shorter generations and better performance. 

%% file: sec_relatedwork.tex
\section{Related Works}
\label{sec:related_work}
\paragraph{Reinforcement Learning for Reasoning Capabilities.}
Recent studies highlight the necessity of reinforcement learning (RL) to unlock complex reasoning capabilities in large language models beyond supervised fine-tuning (SFT)~\cite{shao2024deepseekmath,hou2025t1,yang2025nanbeige43b}. 
~\cite{guo2025deepseek} and ~\cite{wang2025oneshotrlvr} show that pure RL can elicit emergent reasoning behaviors without human-annotated reasoning traces, significantly improving performance on math and STEM abilities. Subsequent studies further improve the effectiveness and practicality of RL for reasoning by expanding RL to real-world software engineering~\cite{pan2024swegym,wei2025swerl} and search~\cite{qi2024webrl,sun2025zerosearch}, pushing the boundaries of RL-driven reasoning in real-world scenarios.

\paragraph{Overthinking in Reasoning Models.} Despite the advanced reasoning capabilities of current reasoning models, recent studies have identified a prevalent ``overthinking'' issue: models often think too much on easy queries~\cite{chen2025donotthinkthatmuch, han-etal-2025-token, yang2025tops}. To mitigate this, existing approaches can be generally categorized into two paradigms: training-free and training-based methods. Training-free methods typically intervene at the inference stage by truncating the model's thinking process to induce a direct answer~\cite{yang2025dynamicearlyexit, wang-etal-2025-wait}. Conversely, training-based methods optimize efficiency by incorporating concise reasoning data or leveraging reinforcement learning. \cite{kang2024c3ot} compresses long rationales into shorter, information-preserving CoTs, then jointly trains an LLM with conditioned long/short CoT formats to get a more efficient model.  Another line of research explicitly incorporate a length penalty into the reinforcement learning process to incentivize shorter sequences~\cite{arora2025trainingreasoningeffectively, kimiteam2025kimik15}. However, such methods often lead to model or length collapse, reducing output length at the expense of model performance.

%% file: sec_method.tex
\section{Methodology: \model} \label{sec:method}
\begin{figure*}[!t]
        \centering
	       \includegraphics[width=1\linewidth]{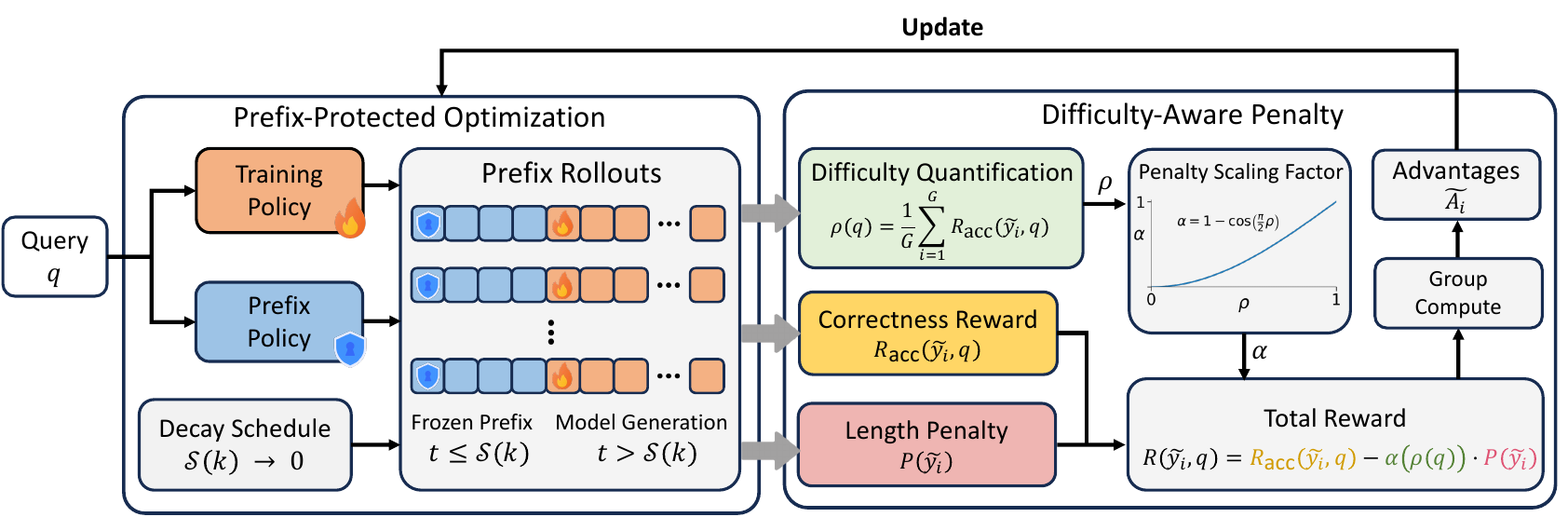}
        \caption{Overview of \model. \model consists of two stages: (1) Prefix-Protected Optimization, which anchors early reasoning by generating a short protected prefix with a frozen prefix policy and gradually decays the prefix length; and (2) Difficulty-Aware Penalty, which scales the length penalty by estimated task difficulty.}
 \label{fig:main}
\end{figure*}

In this section, we detail the PACE: \fullmodel. An overview of PACE is shown in 
Figure~\ref{fig:main}.

\subsection{Prefix-Protected Optimization}\label{sec:prefix_protect}

Existing uniform length penalty methods often force LRMs to over-compress essential initial deduction steps. For example, when applying the uniform length penalty to the arithmetic problem ``$8x+9=23,x=?$'', the compression will start from the first token. However, these early reasoning tokens serve as the logical anchor for the entire solution, and compressing them prematurely disrupts the coherence of the reasoning step.

To mitigate this, we propose the \emph{prefix-protected optimization} that employs a prefix rollout strategy by a frozen prefix policy $\pi_{p}$ to anchor the early generation phase using defined reasoning steps. For the same example ``$8x+9=2,x=?$'', we provide the initial prefix ``To calculate the $x$, we first need to subtract 9...'' as the anchor.
To construct these prefix tokens, we perform rejection sampling~\cite{touvron2023llama} for the initial model on the training set and choose the shortest correct sample as the logical anchor. If no correct sample is available, we use the shortest one. After obtain the prefix, we can generate the entire rollout sequence, take the DeepSeek-R1 template as an example: 
\texttt{\small <|im\_start|>user\textbackslash n 8x+9=23,x=? <|im\_end|>\textbackslash n \allowbreak <|im\_start|>assistant\textbackslash n <think> To calculate the $x$, we first need to subtract 9...then...} 
By locking these initial tokens, we ensure the reasoning step is firmly established before optimization begins.
Specifically, at training step $k$, we construct a hybrid rollout distribution $\pi^{\text{pref}}_{\theta,k}$ that transitions from this frozen policy $\pi_{p}$ (initial model) to the current training policy $\pi_{\theta}$:
\begin{equation}
\small
    \pi^{\text{pref}}_{\theta,k}(y_t \mid y_{<t}, q) = 
    \begin{cases} 
        \pi_{p}(y_t \mid y_{<t}, q) \text{ if } t \le \mathcal{S}(k), \\
        \pi_{\theta}(y_t \mid y_{<t}, q) \text{ if } t > \mathcal{S}(k),
    \end{cases}
    \label{eq:hybrid_policy}
\end{equation}
where $\mathcal{S}(k)$ denotes the prefix length, $\pi_{\theta}$ is the training policy.
By enforcing the prefix $\mathcal{S}(k)$ to follow $ \pi_{p} $, we ensure the optimization starts with a valid logical foundation. 

Since the LRM ultimately needs to solve the task without relying on the frozen policy $ \pi_{p} $, we utilize a linear decay schedule to gradually reduce reliance on the protected policy by progressively shortening the prefix length $\mathcal{S}(k)$ as follows: 

\begin{equation}
\mathcal{S}(k) = 
\begin{cases} 
    \left\lfloor L_{\text{init}} \cdot (1 - \frac{k}{K}) \right\rfloor & \text{if } k < K, \\
    0 & \text{if } k \ge K,
\end{cases}
\label{eq:schedule}
\end{equation}

where $L_{\text{init}}$ denotes the initial prefix length, and $K$ is the number of training steps for which prefix protection is applied.
Initially, the $\mathcal{S}(k)$ strictly constrains the reasoning step.
As training progresses ($k \to K$), this constraint is gradually relaxed, allowing the policy $\pi_{\theta}$ to generate the entire sequence on its own while avoiding over compression.

\subsection{Difficulty-Aware Penalty}\label{sec:reward_scaling}

In real-world reasoning scenarios, the complexity of queries varies significantly, ranging from simple arithmetic to intricate logical deductions.
However, applying a uniform length penalty ignores this critical variance, causing indiscriminative compression at the group level. It penalizes necessary exploration for difficult tasks while failing to curb overthinking on easy ones.
Therefore, to adaptively align length constraints with task difficulty, we propose the \textit{difficulty-aware penalty}.

First, we quantify task difficulties
by leveraging the performance statistics of the generated rollouts.
Specifically, we sample a group of $G$ outputs $\tilde{\mathcal{Y}}^{\text{prefix}}_q = \{\tilde{y}_1, \dots, \tilde{y}_G\}$ from prefix rollout distribution $\pi^{\text{pref}}_{\theta,k}$. Based on these rollouts, we measure the task difficulty through the proxy of the empirical pass rate $\rho(q)$ as follows:
\begin{equation}
    \rho(q) = \frac{1}{G} \sum_{i=1}^G R_{\text{acc}}(\tilde{y}_i, q),
\end{equation}

where $R_{\text{acc}}(\tilde{y}_i, q) = \mathbb{I}\left[ \tilde{y}_i = y^*_q \right]$ is the correctness reward computed by the indicator function $\mathbb{I}[\cdot]$, and $y^*_q$ denotes the ground truth answer. 
Here, $\rho(q)$ serves as an inverse indicator: a lower pass rate implies higher intrinsic difficulty, suggesting that the model struggles to reach the correct solution even with a prefix.

Then, we formulate the total reward $R(\tilde{y}_i, q)$ using adaptive penalty modulation. 
We couple the correctness reward with a dynamic length penalty, establishing a dependency between the penalty intensity and the pass rate $\rho(q)$:
\begin{equation}
    R(\tilde{y}_i, q) = R_{\text{acc}}(\tilde{y}_i, q) - \alpha \big(\rho(q) \big) \cdot P(\tilde{y}_i),
\end{equation}
where $P(\tilde{y}_i) = \frac{L(\tilde{y}_i)-L_{min}}{L_{max}-L_{min}}$ is the min–max length penalty~\cite{kimiteam2025kimik15}. We observe similar trends on the other length penalties shown in Section~\ref{sec: generalization}. $\alpha\big(\rho(q)\big)$ is the scaling factor and defined as follows:
\begin{equation}
    \alpha\big(\rho(q)\big) = 1 - \cos\left(\frac{\pi}{2}\rho(q)\right).
\end{equation}
Here, for simple problems where $\rho(q) \to 1$, the scaling factor $\alpha$ approaches its maximum, strictly enforcing conciseness to eliminate redundancy. 
Conversely, for highly complex problems where $\rho(q) \to 0$, thus $\alpha$ decays towards zero. 
This effectively shields the model from length penalties on complex tasks, granting it the necessary computational budget to perform deep reasoning without incurring negative rewards.
 
Finally, we integrate the above two components into the GRPO framework~\cite{shao2024deepseekmath}. We compute the advantages by standardizing the difficulty-aware rewards within the generated group $\tilde{\mathcal{Y}}^{\text{prefix}}_q$ as follows:
\begin{equation}
    \tilde{A}_i = \frac{\tilde{R}(\tilde{y}_i,q) - \tilde{\mu}_{\text{q}}}{\tilde{\sigma}_{\text{q}} + \epsilon},
\end{equation}
where $\tilde{\mu}_{\text{q}}$ and $\tilde{\sigma}_{\text{q}}$ denote the mean and standard deviation of the total reward, $\epsilon$ is a small constant introduced to prevent division by zero.

Thus, the policy parameters $\theta$ are updated by maximizing the following objective:
\begin{equation}
\small
\begin{aligned}
\mathcal{J}(\theta)
= \mathbb{E}
\Big[ \min \big( \gamma_i \tilde{A}_i,\; \operatorname{clip}(\gamma_i, 1-\varepsilon, 1+\varepsilon)\, \tilde{A}_i \big) \\
\qquad\qquad -\; \beta \, \mathbb{D}_{\text{KL}}\!\left(\pi^{\text{pref}}_{\theta,k}(\cdot \mid ,q)\;\|\;\pi_{\text{ref}}(\cdot \mid,q)\right) \Big],
\end{aligned}
\end{equation}
where $\gamma_i$ is the importance sampling ratio and $\beta$ is the coefficient that weights the KL-divergence penalty.
This objective effectively combines sequence-level prefix protection with group-level adaptive penalty, driving the model to achieve high efficiency by eliminating redundancy without compromising reasoning accuracy.

%% file: sec_experiment.tex
\section{Experimental Setup}

\paragraph{Datasets and Metrics}
We initialize our models from DeepSeek-R1-Distill-Qwen-7B and DeepSeek-R1-Distill-Qwen-1.5B. 
We choose the widely used reinforcement learning dataset Skywork-o1~\cite{he2025skywork} as our training set. 
We evaluate the performance of our proposed method on in-domain and out-of-domain tasks.
For in-domain tasks, we choose four challenging mathematical benchmarks: AIME24, AIME25~\cite{aime_1983_2024}, MATH500~\cite{hendrycks2021measuring}, and GSM8K~\cite{pan2024training}. 
For out-of-domain tasks, we choose three representative benchmarks on different domains: science~(GPQA-D~\cite{rein2024gpqa}), code~(LiveCodeBench-v6~\cite{jain2024livecodebench}), and instruction following~(IF Eval~\cite{zhou2023instruction})
We report three primary metrics: 
(1) Average Accuracy (\textbf{ACC}, \%); 
(2) Average Token Usage (\textbf{LEN}); and 
(3) Token Reduction Rate (\textbf{TR}), defined as the percentage reduction in average output length relative to the Vanilla model, which quantifies the efficiency gain in terms of computational cost. 
Without specifying, we report Pass@1 accuracy averaged over 16 independent runs per prompt, using a context window of 32k and a sampling temperature of 0.6. All evaluation details can be found in Appendix~\ref{app:benchmarks}.

\paragraph{Baselines}
We compare our PACE against two categories of representative baselines:

\noindent
$\bullet$ \textbf{Training-free Methods:} These methods typically employ inference-time intervention. DEER~\cite{yang2025dynamicearlyexit}, Dynasor-CoT~\cite{fu2025reasoning} implement dynamic early exit by monitoring intermediate confidence/certainty signals during generation and stopping CoT once a reliable answer is detected. Thinkswitcher~\cite{liang2025thinkswitcher} utilizes a trained switcher module to dynamically toggle the generation of reasoning chains, deciding whether to ``think'' before answering.

\noindent
$\bullet$ \textbf{Training-based Methods:} These methods integrate efficiency constraints during training. Efficient-R~\cite{arora2025trainingreasoningeffectively} applies a uniform length penalty as a reward signal to all outputs. O1-Pruner~\cite{luo2025o1} harmonizes output length by pre-sampling to estimate a reference/baseline behavior, and then applies RL-style optimization to penalize redundant long reasoning while constraining accuracy. AdaptThink~\cite{zhang2025adaptthink} trains a reasoning model with RL to adaptively switch between Thinking and NoThinking modes based on input difficulty. DAST~\cite{shen2025dast} quantifies difficulty via a token-length budget and applies budget-aware reward shaping with preference optimization to regulate per-query CoT length, reducing overthinking without sacrificing accuracy. Dr.SAF~\cite{chen2025aware} facilitates dynamic termination of the chain-of-thought process via boundary-awareness alignment followed by reinforcement learning.

\paragraph{Implementation Details}
We utilize verl~\cite{sheng2024hybridflow} as the reinforcement learning framework on 32 NVIDIA A100 GPUs and set the batch size as 64 with a group size as 8. We choose $L_{\text{init}}$ to be 512 and apply prefix protection with $K$ set to 100. We also set the learning rate to $1\times 10^{-6}$ with 10 warm-up steps. More training details can be found in Appendix~\ref{sec: training parameter}.

\input{tables/main_results}

\section{Experimental Results}
\subsection{Main Results}

\input{tables/ood_eval}

\paragraph{\model breaks the efficiency-performance trade-off.} 
As shown in Table~\ref{tab:main_results}, \model shows superior performance in terms of both token efficiency and accuracy across benchmarks and model sizes. 
Notably, the most significant observation is that \model is the only method that achieves a substantial reduction in token usage ($>45\%$) while simultaneously \textit{improving} average accuracy over the Vanilla baseline. 
In contrast, existing methods struggle to balance this trade-off: aggressive pruners like DR.SAF and DEER achieve high token reduction rates (TR) but suffer from evident performance degradation (e.g., DR.SAF drops accuracy by 2.7\% and DEER drops accuracy by 2.3\% on 7B model). 
Conversely, conservative methods like AdaptThink preserve accuracy but yield limited efficiency gains (only 28.5\% TR on 7B).
However, \model dominates the Pareto frontier. 
On the 7B model, we achieve a \textbf{55.7\%} token reduction, which is nearly double the efficiency of the second-best performer, AdaptThink, while surpassing the vanilla model's accuracy by \textbf{0.6\%}.
On the smaller 1.5B model, the gain is even more pronounced: \model cuts the average length with 46\% token reduction while boosting average accuracy from \textbf{53.2} to \textbf{57.3} (+4.1), indicating that \model pushes the Pareto frontier of reasoning compression.

\paragraph{\model also shows great generalization to other domains} Table~\ref{tab:ood_eval} presents the performance of our \model on out-of-domain (OOD) tasks. Although trained exclusively on mathematical datasets, our \model exhibits consistent gains in both accuracy and token efficiency across science (GPQA-D), code (LiveCodeBench-v6), and instruction following (IF Eval). 
Specifically, the 1.5B model surpasses the vanilla baseline with 1.1\%--4.6\% accuracy improvements while reducing generation length by 14.5\%--37.0\%. 
A similar trend is observed in the 7B model, which achieves up to 1.8\% accuracy gain with significant token reductions ranging from 25.7\% to 33.3\%.  
These results yield two key insights: (1) Beyond the mathematical domain, other tasks (\eg code) also have the potential to shorten their reasoning chains while maintaining or improving the performance. 
(2) Gains in both performance and token reduction are consistently enjoyed across domains, further showcasing the effectiveness and robustness of our proposed dual-level control method.

\begin{figure*}[!t]
        \centering
	\includegraphics[width=1\linewidth]{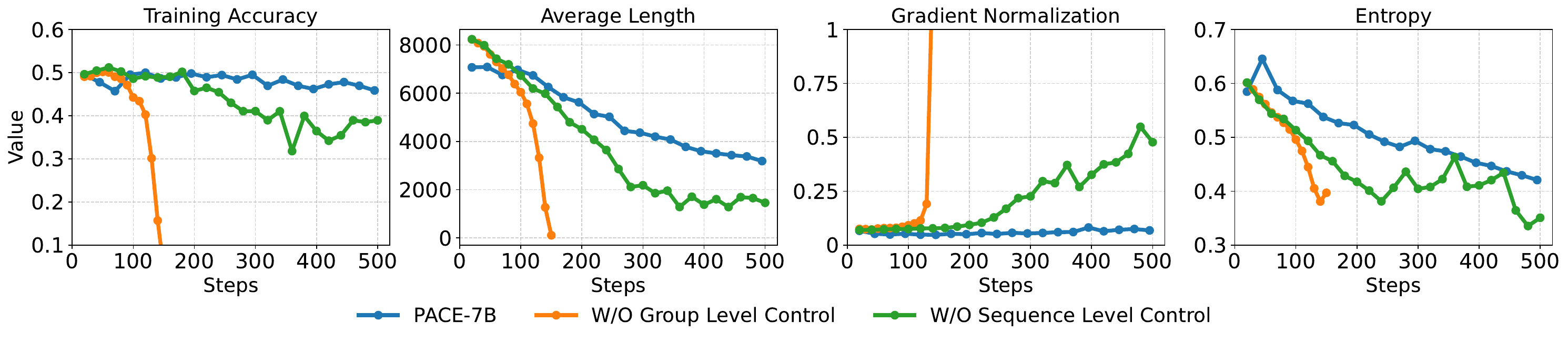}
        \caption{Training dynamics of the PACE-7B and its ablations over RL steps, reporting training accuracy, average response length, gradient normalization, and policy entropy.}
 \label{fig:exp_training}
\end{figure*}
\input{tables/ablation}

\subsection{Ablation Study}
\paragraph{Dual-level control is crucial for performance and token efficiency} To investigate the efficacy of our proposed dual-level control mechanism, we conducted an ablation study by training a 7B model without sequence-level and group-level controls under identical settings. 
As shown in Table~\ref{tab:ablation_study}, removing group-level control significantly harms the model's internal reasoning ability, causing 9.8\% performance drop on AIME24 and 12.9\% on MATH500. 
We attribute this deterioration to indiscriminative compression of questions, causing the model to suffer from reasoning collapse during training. 

Removing sequence-level control also degrades performance sharply, as compression starts from the initial tokens, leading the model to skip essential reasoning steps. 
In contrast, sequence-level control effectively guides the policy model by validating solution entry points, ensuring that core reasoning steps are preserved during compression.

\paragraph{Dual-level control stabilizes training process} 

We also present the 7B model ablation training dynamics across four metrics: average length, gradient norm, training accuracy, and entropy. 
As shown in Figure~\ref{fig:exp_training}, dual-level control sustains training accuracy while the response length decreases. 
Without group-level or sequence-level control, the model often sacrifices training accuracy for length reduction. 
Moreover, dual-level control can stabilize the training process, exhibiting consistent gradient norms and entropy. 

Our gradient norm remains stable throughout, unlike the explosion in the other two ablations.
We also observe benefits in entropy evolution, especially at the later stages of training. 
Our method significantly prevents entropy oscillation, indicating that the model retains steady reasoning capabilities.
\begin{figure}[!t]
        \centering
	\includegraphics[width=0.6\linewidth]{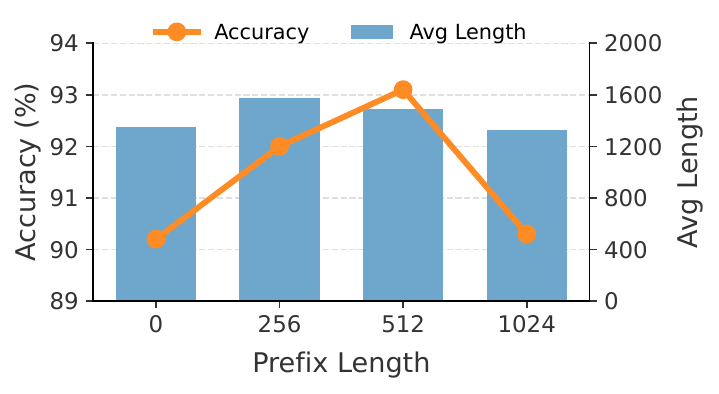}
        \caption{Impact of different prefix length on MATH500 dataset.}
 \label{fig:exp_prefix_length}
\end{figure}

\begin{figure}[t]
    \centering
    \begin{minipage}[t]{0.49\linewidth}
        \centering
        \includegraphics[width=\linewidth]{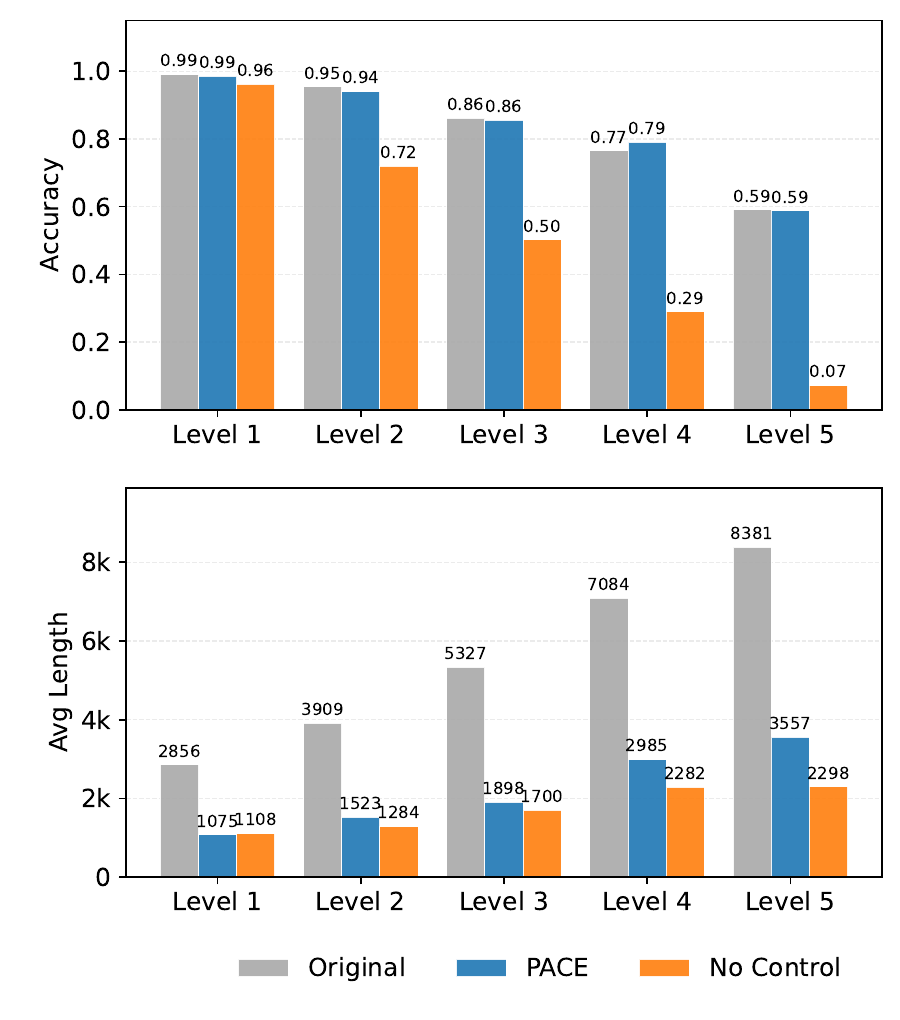}
    \end{minipage}\hfill
    \begin{minipage}[t]{0.49\linewidth}
        \centering
        \includegraphics[width=\linewidth]{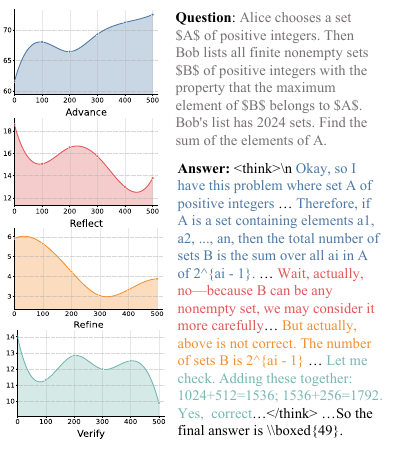}
    \end{minipage}

    \caption{Left: Accuracy and average length change across different difficulty levels on MATH500 dataset. Right: The distributional shift of four reasoning behaviors (Advance, Reflect, Refine, Verify) across training steps, with a sample AIME 2024 solution color-coded by behavior. Detailed prompt and explanation can be seen on Appendix \ref{app:prompt}.}
    \label{fig:combined_difficulty_distribution}
\end{figure}

\subsection{Prefix Length Study}

To study the effect of sequence-level prefix length, we trained the 7B model with varying prefix length $L_{\text{init}}$ (0, 256, 512, and 1024 tokens) under identical settings and evaluated on the MATH500 dataset.
As shown in Figure~\ref{fig:exp_prefix_length}, sequence-level prefix protection clearly improves performance. Even with a shorter prefix (256 tokens), accuracy rises from 90.2\% to 92.0\% by preserving valuable reasoning steps. This improvement continues at 512 tokens, where the model achieves a peak accuracy of 93.1\%. However, an excessively long prefix (1024 tokens) harms performance, dropping accuracy to 90.3\%. This is because during the early training stages, shorter samples are fully encapsulated within the 1024-token prefix, exposing the ground truth answer immediately and enabling the model to bypass reasoning steps. Consequently, the model exhibits ``shortcut learning'' and degrades in reasoning capability. 

\subsection{Across Difficulty Levels Study}

The core objective of \model is to compress without sacrificing the model's underlying reasoning capability. We evaluate it against the original model and a ``No Control'' baseline across difficulty levels on MATH500 dataset. 

As shown in Figure~\ref{fig:combined_difficulty_distribution} left, in the more complex categories (Levels 3-5), where multi-step reasoning is paramount, \model's performance remains comparable to the original model while requiring significantly fewer tokens. At Level 4, \model even slightly outperforms the original model (0.79 vs 0.77) with a 58\% token reduction, indicating that targeted compression can help reduce redundant noise in complex reasoning steps. 

In contrast, the ``No Control'' variant suffers catastrophic performance collapse as difficulty increases, with accuracy dropping from 0.96 (Level 1) to just 0.07 (Level 5),
reflecting degraded reasoning ability due to excessive compression on critical reasoning steps. 
This significant gap highlight that without dual-level control, the model tends to ``over-compress'' essential logic, whereas \model can preserve core reasoning while achieving conciseness.

\label{sec: generalization}
\begin{figure}[!t]
        \centering
	\includegraphics[width=0.8\linewidth]{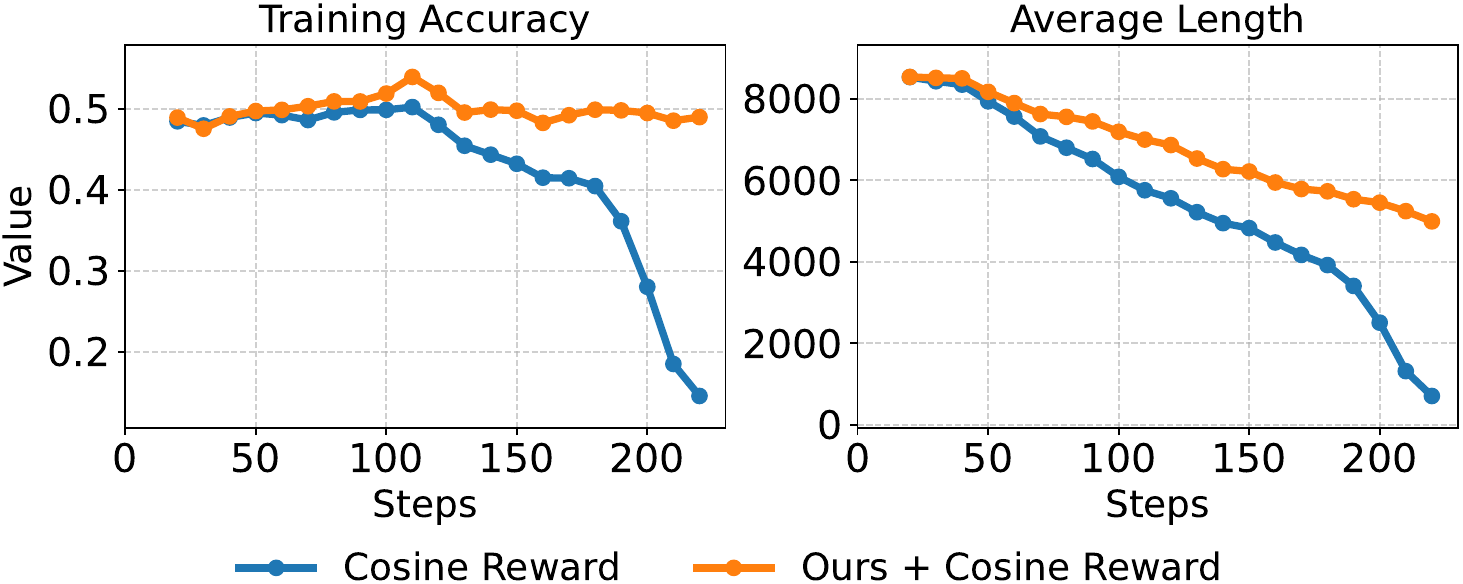}
        \caption{Training dynamics of a cosine reward compared to our \model.}
 \label{fig:other_reward}
\end{figure}

\input{tables/other_rewards}
\subsection{Reasoning Behavior Shift During Training}
Although PACE significantly reduces token usage, a natural concern arises: \textit{Does the model simply learn to skip crucial reasoning steps rather than pruning unnecessary ones?} To investigate which part of PACE reduces token usage and how each part changes during training, we define four reasoning behaviors for reasoning models: Advance, Reflect, Refine and Verify, as color-coded in Figure~\ref{fig:combined_difficulty_distribution} right. We test checkpoints during training with 100 step interval on AIME2024\&25 across 16 independent runs. Using Qwen3-32B~\cite{yang2025qwen3} to annotate the generated chains of thought, we calculated 
the proportion of each behavior. 

As shown in Figure~\ref{fig:combined_difficulty_distribution} right, a clear distributional shift emerges: the Advance behavior steadily rise from 61.5\% to 72.5\%, 

while Reflect and Verify decline from 18.5\% to 13.6\% and 14.0\% to 9.8\%, respectively. 
This indicates that token reduction is primarily comes from suppressing redundant self-correction and excessive hesitation, rather than omitting constructive logic. During training, the model is encouraged to ``think forward'' more confidently, resulting in a higher ratio of AAdvance steps. Consequently, PACE promotes more efficient reasoning by emphasizing decisive problem-solving steps over repetitive verification loops.

\subsection{Generalization to Other Length Rewards}
To demonstrate the generalization ability and robustness of \model, we extended our evaluation to include alternative reward mechanisms. 
Following prior work\cite{arora2025trainingreasoningeffectively}, we adopted a static cosine reward function to penalize generation length, which is formulated as:
\begin{equation}
    P(\tilde{y}_i) = \sigma \left( \frac{L(\tilde{y}_i) - L_{mean} } { L_{std}  } \right),
\end{equation}
where $L(\tilde{y}_i)$ is the length of the $i$-th rollout $\tilde{y}_i$, while $L_{mean}$ and $ L_{std}$ are the mean and standard deviation of lengths for all rollouts for query $q$ and $\sigma$ is the sigmoid activation function.

Figure~\ref{fig:other_reward} shows the training dynamics of the baseline static cosine reward (blue) compared to our method (orange). The baseline exhibits a critical instability: 

as it only optimizes for the length reward, the training accuracy (Left) suffers a catastrophic collapse after approximately 150 steps.
The length plot (Right) confirms that the baseline aggressively shortens outputs, compromising reasoning capabilities and yielding an unacceptable efficiency-performance trade-off.

Table~\ref{tab:other_rewards} further shows that these superior training dynamics directly contribute to stronger final performance. PACE consistently outperforms the Cosine Reward baseline across all benchmarks, demonstrating robust generalization to different reward formulations and a more stable, superior Pareto frontier between inference cost and reasoning performance.

%% file: tables/main_results.tex
\newcommand{\acc}[2]{#1\kern0.15em\raisebox{-0.65ex}{\scriptsize #2}}
\newcommand{\accpos}[1]{\textcolor{green!60!black}{+{#1}}}
\newcommand{\accneg}[1]{\textcolor{red}{-{#1}}}
\definecolor{TrainFree}{HTML}{D7ECFF}   
\definecolor{TrainBased}{HTML}{E8DDF8}  

\DeclareRobustCommand{\capbox}[2]{%
  \begingroup
  \setlength{\fboxsep}{0.25pt}%
  \setlength{\fboxrule}{0pt}%
  \colorbox{#1}{\raisebox{0pt}[0.85em][0.15em]{#2}}%
  \endgroup
}

\begin{table*}[!t]
\centering
\resizebox{\textwidth}{!}{%
\begin{tabular}{lccccccccccccccc}
\toprule
\multirow{2}{*}{\textbf{Method}} & \multicolumn{3}{c}{\textbf{AIME24}} & \multicolumn{3}{c}{\textbf{AIME25}} & \multicolumn{3}{c}{\textbf{MATH500}} & \multicolumn{3}{c}{\textbf{GSM8k}} & \multicolumn{3}{c}{\textbf{AVG}} \\
\cmidrule(lr){2-4} \cmidrule(lr){5-7} \cmidrule(lr){8-10} \cmidrule(lr){11-13} \cmidrule(lr){14-16}
 & Acc$\uparrow$ & Len$\downarrow$ & TR$\uparrow$ & Acc$\uparrow$ & Len$\downarrow$ & TR$\uparrow$ & Acc$\uparrow$ & Len$\downarrow$ & TR$\uparrow$ & Acc$\uparrow$ & Len$\downarrow$ & TR$\uparrow$ & Acc$\uparrow$ & Len$\downarrow$ & TR$\uparrow$ \\
\midrule
\multicolumn{16}{c}{\textit{\textbf{DeepSeek-R1-Distill-Qwen-7B}}} \\
\midrule
Vanilla &
\acc{52.7}{\phantom{-10}} & 12228 & - &
\acc{38.6}{\phantom{-10}} & 12977 & - &
\acc{92.8}{\phantom{-10}} & 3832 & - &
\acc{91.5}{\phantom{-10}} & 687 & - &
\acc{68.9}{\phantom{-10}} & 7431 & - \\

\rowcolor{TrainFree} Thinkswitcher &
\acc{48.3}{\accneg{4.4}} & 7936 & 35.1\% &
\acc{37.5}{\accneg{1.1}} & 6955 & 46.4\% &
\acc{91.3}{\accneg{1.5}} & 3495 & 8.8\% &
\acc{\underline{92.5}}{\accpos{1.0}} & 1389 & -102.2\% &
\acc{67.4}{\accneg{1.5}} & 4944 & 33.5\% \\

\rowcolor{TrainFree} DEER &
\acc{49.2}{\accneg{3.5}} & 9839 & 19.5\% &
\acc{36.7}{\accneg{1.9}} & 7257 & 44.1\% &
\acc{89.8}{\accneg{3.0}} & 2143 & 44.1\% &
\acc{90.6}{\accneg{0.9}} & 917 & -33.5\% &
\acc{66.6}{\accneg{2.3}} & 5039 & 32.2\% \\

\rowcolor{TrainFree} Dynasor-CoT &
\acc{46.7}{\accneg{6.0}} & 12695 & -3.8\% &
\acc{33.3}{\accneg{5.3}} & 11069 & 14.7\% &
\acc{89.0}{\accneg{3.8}} & 2971 & 22.5\% &
\acc{89.6}{\accneg{1.9}} & 1285 & -87.0\% &
\acc{64.7}{\accneg{4.2}} & 7005 & 5.7\% \\

\rowcolor{TrainBased} O1-Pruner &
\acc{49.2}{\accneg{3.5}} & 9719 & 20.5\% &
\acc{35.0}{\accneg{3.6}} & 8236 & 36.5\% &
\acc{91.5}{\accneg{1.3}} & 2268 & 40.8\% &
\acc{91.1}{\accneg{0.4}} & 1012 & -47.3\% &
\acc{66.7}{\accneg{2.2}} & 5309 & 28.6\% \\

\rowcolor{TrainBased} DAST &
\acc{45.6}{\accneg{7.1}} & 7578 & 38.0\% &
\acc{37.0}{\accneg{1.6}} & 10729 & 17.3\% &
\acc{89.6}{\accneg{3.2}} & 2162 & 43.6\% &
\acc{86.7}{\accneg{4.8}} & 489 & 28.8\% &
\acc{64.7}{\accneg{4.2}} & 5240 & 29.5\% \\

\rowcolor{TrainBased} AdaptThink &
\acc{\textbf{52.9}}{\accpos{0.2}} & 9596 & 21.5\% &
\acc{38.3}{\accneg{0.3}} & 10670 & 17.8\% &
\acc{92.0}{\accneg{0.8}} & 1875 & 51.1\% &
\acc{90.3}{\accneg{1.2}} & 408 & 40.6\% &
\acc{\underline{69.1}}{\accpos{0.2}} & 5316 & 28.5\% \\

\rowcolor{TrainBased} DR.SAF &
\acc{50.6}{\accneg{2.1}} & \underline{6288} & \underline{48.6\%} &
\acc{38.2}{\accneg{0.4}} & \underline{6747} & \underline{48.0\%} &
\acc{88.3}{\accneg{4.5}} & \textbf{1061} & \textbf{72.3\%} &
\acc{88.1}{\accneg{3.4}} & \textbf{162} & \textbf{76.4\%} &
\acc{66.3}{\accneg{2.6}} & \underline{3565} & \underline{52.0\%} \\

\rowcolor{TrainBased} Efficient-R &
\acc{51.9}{\accneg{0.8}} & 7464 & 39.0\% &
\acc{33.3}{\accneg{5.3}} & 8902 & 31.4\% &
\acc{88.5}{\accneg{4.3}} & \underline{1306} & \underline{65.9\%} &
\acc{87.2}{\accneg{4.3}} & \underline{263} & \underline{61.7\%} &
\acc{65.2}{\accneg{3.7}} & 4484 & 39.7\% \\

\rowcolor{TrainBased} \textbf{\model} &
\acc{\underline{52.7}}{\accpos{0.0}} & \textbf{5525} & \textbf{54.8\%} &
\acc{\textbf{39.4}}{\accpos{0.8}} & \textbf{5849} & \textbf{54.9\%} &
\acc{\textbf{93.1}}{\accpos{0.3}} & 1490 & 61.1\% &
\acc{\textbf{92.6}}{\accpos{1.1}} & 309 & 55.0\% &
\acc{\textbf{69.5}}{\accpos{0.6}} & \textbf{3293} & \textbf{55.7\%} \\

\midrule
\multicolumn{16}{c}{\textit{\textbf{DeepSeek-R1-Distill-Qwen-1.5B}}} \\
\midrule
Vanilla &
\acc{28.3}{\phantom{-10}} & 12094 & - &
\acc{25.0}{\phantom{-10}} & 11860 & - &
\acc{84.0}{\phantom{-10}} & 4627 & - &
\acc{75.5}{\phantom{-10}} & 987 & - &
\acc{53.2}{\phantom{-10}} & 7392 & - \\

\rowcolor{TrainFree} Thinkswitcher &
\acc{23.3}{\accneg{5.0}} & 8192 & 32.3\% &
\acc{\underline{28.3}}{\accpos{3.3}} & \underline{6689} & \underline{43.6\%} &
\acc{82.4}{\accneg{1.6}} & 4544 & 1.8\% &
\acc{\textbf{84.7}}{\accpos{9.2}} & 2114 & -114.2\% &
\acc{\underline{54.7}}{\accpos{1.5}} & 5385 & 27.2\% \\

\rowcolor{TrainFree} DEER &
\acc{23.3}{\accneg{5.0}} & 9557 & 21.0\% &
\acc{10.0}{\accneg{15.0}} & 9281 & 21.7\% &
\acc{67.8}{\accneg{16.2}} & 2497 & 46.0\% &
\acc{74.7}{\accneg{0.8}} & 984 & 0.3\% &
\acc{44.0}{\accneg{9.2}} & 5580 & 24.5\% \\

\rowcolor{TrainBased} O1-Pruner &
\acc{28.9}{\accpos{0.6}} & 10361 & 14.3\% &
\acc{23.2}{\accneg{1.8}} & 8731 & 26.4\% &
\acc{\underline{82.2}}{\accneg{1.8}} & 3212 & 30.6\% &
\acc{74.8}{\accneg{0.7}} & 458 & 53.6\% &
\acc{52.3}{\accneg{0.9}} & 5691 & 23.0\% \\

\rowcolor{TrainBased} AdaptThink &
\acc{\underline{30.8}}{\accpos{2.5}} & 7616 & 37.0\% &
\acc{25.8}{\accpos{0.8}} & 7456 & 37.1\% &
\acc{82.0}{\accneg{2.0}} & \textbf{1685} & \textbf{63.6\%} &
\acc{\underline{81.2}}{\accpos{5.7}} & 439 & 55.5\% &
\acc{55.0}{\accpos{1.8}} & \underline{4299} & \underline{41.8\%} \\

\rowcolor{TrainBased} Efficient-R &
\acc{24.6}{\accneg{3.7}} & 6945 & \underline{42.6\%} &
- & - & - &
\acc{79.0}{\accneg{5.0}} & 2105 & 54.5\% &
\acc{78.4}{\accpos{2.9}} & \textbf{300} & \textbf{69.6\%} &
- & - & - \\

\rowcolor{TrainBased} \textbf{\model} &
\acc{\textbf{33.3}}{\accpos{5.0}} & \textbf{6840} & \textbf{43.4\%} &
\acc{\textbf{29.6}}{\accpos{4.6}} & \textbf{6592} & \textbf{44.4\%} &
\acc{\textbf{85.8}}{\accpos{1.8}} & 1987 & \underline{57.1\%} &
\acc{80.3}{\accpos{4.8}} & \underline{357} & \underline{63.8\%} &
\acc{\textbf{57.3}}{\accpos{4.1}} & \textbf{3944} & \textbf{46.6\%} \\
\bottomrule
\end{tabular}%
}
\caption{Performance comparison on AIME24, AIME25, MATH500, and GSM8K datasets. Rows shaded in \capbox{TrainFree}{lavender} indicate training-free (inference-time) methods, while rows shaded in \capbox{TrainBased}{light purple} indicate training-based methods.
Best results are highlighted in \textbf{bold}, second best results are highlighted in \underline{underline}.}
\label{tab:main_results}
\end{table*}

%% file: tables/ood_eval.tex
\definecolor{TrainBased}{HTML}{E8DDF8}  

\begin{table}[!t]
\centering

\begin{tabular}{lccccccccc}
\toprule
\multirow{2}{*}{\textbf{Method}} & \multicolumn{3}{c}{\textbf{GPQA-D}} & \multicolumn{3}{c}{\textbf{LiveCodeBench}} & \multicolumn{3}{c}{\textbf{IF Eval}} \\
\cmidrule(lr){2-4} \cmidrule(lr){5-7} \cmidrule(lr){8-10}
 & Acc$\uparrow$ & Len$\downarrow$ & TR$\uparrow$ & Acc$\uparrow$ & Len$\downarrow$ & TR$\uparrow$ & Acc$\uparrow$ & Len$\downarrow$ & TR$\uparrow$ \\
\midrule
\multicolumn{10}{c}{\textit{\textbf{DeepSeek-R1-Distill-Qwen-7B}}} \\
\midrule
Vanilla & 49.3 & 8011 & - & 35.0 & 11723 & - & 60.2 & 1570 & - \\
\rowcolor{TrainBased} \textbf{\model} & \textbf{49.9} & \textbf{5346} & \textbf{33.3\%} & \textbf{36.8} & \textbf{8110} & \textbf{30.8\%} & \textbf{60.8} & \textbf{1167} & \textbf{25.7\%} \\
\midrule
\multicolumn{10}{c}{\textit{\textbf{DeepSeek-R1-Distill-Qwen-1.5B}}} \\
\midrule
Vanilla & 36.3 & 9342 & - & 17.2 & 13991 & - & 40.7 & 5249 & - \\
\rowcolor{TrainBased} \textbf{\model} & \textbf{39.4} & \textbf{5889} & \textbf{37.0\%} & \textbf{18.3} & \textbf{10286} & \textbf{26.5\%} & \textbf{45.3} & \textbf{4487} & \textbf{14.5\%} \\
\bottomrule
\end{tabular}%

\vspace{10pt} 
\caption{Results on out of domain (OOD) evaluation across science (GPQA-D), code (LiveCodeBench-v6), and instruction following (IF Eval).}
\label{tab:ood_eval}
\vspace{-15pt} 
\end{table}

%% file: tables/ablation.tex
\definecolor{TrainBased}{HTML}{E8DDF8}

\begin{table}[!t]
\centering

\begin{tabular}{lcccccc}
\toprule
\multirow{2}{*}{\textbf{Method}} &
\multicolumn{2}{c}{\textbf{AIME24}} &
\multicolumn{2}{c}{\textbf{AIME25}} &
\multicolumn{2}{c}{\textbf{MATH500}} \\
\cmidrule(lr){2-3} \cmidrule(lr){4-5} \cmidrule(lr){6-7}
 & Acc$\uparrow$ & Len$\downarrow$ & Acc$\uparrow$ & Len$\downarrow$ & Acc$\uparrow$ & Len$\downarrow$ \\
\midrule
\rowcolor{TrainBased} \textbf{\model} & \textbf{52.7} & \textbf{5525} & \textbf{39.4} & \textbf{5849} & \textbf{93.1} & \textbf{1390} \\
\quad w/o group-level control    & 42.5 & 5396 & 31.6 & 6538 & 80.2 & 1414 \\
\quad w/o sequence-level control & 45.6 & 4952 & 32.5 & 4887 & 90.2 & 1351 \\
\bottomrule
\end{tabular}%

\vspace{10pt}
\caption{Ablation study on mathematical benchmarks.}
\label{tab:ablation_study}
\vspace{-15pt}
\end{table}

%% file: tables/other_rewards.tex
\definecolor{TrainBased}{HTML}{E8DDF8}

\begin{table}[!t]
\centering
\begin{tabular}{lcccccc}
\toprule
\multirow{2}{*}{\textbf{Method}} &
\multicolumn{2}{c}{\textbf{AIME24}} &
\multicolumn{2}{c}{\textbf{AIME25}} &
\multicolumn{2}{c}{\textbf{MATH500}} \\
\cmidrule(lr){2-3} \cmidrule(lr){4-5} \cmidrule(lr){6-7}
 & Acc$\uparrow$ & Len$\downarrow$ & Acc$\uparrow$ & Len$\downarrow$ & Acc$\uparrow$ & Len$\downarrow$ \\
\midrule
Cosine Reward & 47.0 & 6714 & 34.1 & 7817 & 88.8 & 2364 \\
\rowcolor{TrainBased} \textbf{\model} & \textbf{51.7} & \textbf{6873} & \textbf{38.8} & \textbf{7517} & \textbf{91.4} & \textbf{2035} \\
\bottomrule
\end{tabular}%
\vspace{10pt}
\caption{Performance comparison between other length reward (\eg cosine reward) and our \model.}
\label{tab:other_rewards}
\vspace{-15pt}
\end{table}

%% file: sec_conclusion.tex
\section{Conclusion}
We propose \model, a dual-level framework for length compression of LRMs. \model consists of a prefix-protected optimization to preserve essential early reasoning and a difficulty-aware length penalty to avoid disproportionately compressing on complex tasks. Experimental results show that \model significantly reduces inference costs and further improves accuracy across multiple domains. Moreover, \model generalizes to alternative length-reward designs, underscoring the effectiveness of dual-level protection in pushing the efficiency–accuracy Pareto frontier.

\section*{Limitations}
Beyond the strong accuracy–efficiency trade-off achieved by \model, there remains substantial headroom to further improve efficiency, as roughly 40\% of tokens are still devoted to reasoning. In addition, while we observe promising out-of-domain performance on several benchmarks, a broader and more systematic study of generalization is needed, covering more diverse domains, prompt and format variations, and a wider range of distribution shifts to better understand when and why adaptive compression transfers.

%% file: sec_appendix.tex
\section{Evaluation Tasks Details}
\label{app:benchmarks}
We report pass@1 scores based on 16 independent runs for AIME24, AIME25, MATH500, GSM8K, and GPQA-D, and 4 independent runs for LiveCodeBench-v6. For IFEval, we evaluate using pass@1 directly due to its extensive size. All experiments were conducted using a 32k context window. Following is the details and resources for each tasks.

\noindent\textbf{In-domain mathematical reasoning benchmarks.}
\begin{itemize}
    \item \textbf{AIME24}
    We use the 2024 American Invitational Mathematics Examination problems (AIME I \& II; 30 problems in total), where each question requires multi-step contest-level reasoning and the final answer is a short integer. We report accuracy via exact-match of the final answer. \url{https://huggingface.co/datasets/HuggingFaceH4/aime_2024}

    \item \textbf{AIME25}
    Similar to AIME24, this benchmark contains the 2025 AIME I \& II problems (30 problems total) and tests olympiad-style mathematical reasoning with exact-match final-answer evaluation. \url{https://huggingface.co/datasets/math-ai/aime25}

    \item \textbf{MATH500}
    MATH500 is a curated 500-problem subset derived from the MATH competition dataset, covering diverse topics (e.g., algebra, geometry, number theory) and emphasizing solution correctness on challenging problems. \url{https://huggingface.co/datasets/HuggingFaceH4/MATH-500}

    \item \textbf{GSM8K}
    GSM8K is a grade-school math word-problem benchmark designed for multi-step arithmetic reasoning, with reference solutions and final numeric answers. We use its test set (1319 samples) for our evaluation. \url{https://huggingface.co/datasets/openai/gsm8k}
\end{itemize}

\noindent\textbf{Out-of-domain (OOD) generalization benchmarks.}
\begin{itemize}
    \item \textbf{GPQA-D (Science)}
    GPQA is a graduate-level, “Google-proof” multiple-choice QA benchmark written by domain experts across biology, chemistry, and physics. We use its Diamond (GPQA-D, 198 questions) split as a challenging science OOD evaluation and report multiple-choice accuracy. \url{https://huggingface.co/datasets/Idavidrein/gpqa}

    \item \textbf{LiveCodeBench-v6 (Code)}
    LiveCodeBench is a contamination-aware, continuously updated coding benchmark built from newly released contest problems, and supports programmatic grading via unit tests. We use the latest v6 release for our evaluation. \url{https://github.com/LiveCodeBench/LiveCodeBench}

    \item \textbf{IFEval (Instruction Following).}
    IFEval evaluates instruction-following using programmatically verifiable constraints (e.g., required keywords, length constraints, formatting rules), enabling reproducible automatic checking. It contains 25 instruction types and 541 prompts; we report instruction-following accuracy under the benchmark’s verifier. \url{https://github.com/google-research/google-research/tree/master/instruction_following_eval}
\end{itemize}

\section{Effectiveness of Prefix}
In this section, we demonstrate the effectiveness of prefix protection. Concretely, we construct the prefix via rejection sampling on AIME 2024 and AIME 2025 with DeepSeek-R1-Distill-Qwen-7B, selecting the shortest correct sample for each problem. When no correct sample is available, we fall back to the shortest sampled response. We then vary the protected prefix length and report pass@1 accuracy averaged over 8 independent runs. As shown in Table~\ref{tab:prefix_aime}, prefix protection yields a clear accuracy--efficiency improvement. Compared to using no prefix (0), a 600-token prefix improves AIME24 accuracy from 52.7 to 57.8 ($+5.1$ points) while reducing generation from 12,228 to 8,568 tokens ($-3,660$). On AIME25, longer anchors are particularly beneficial: with a 1,000-token prefix, accuracy increases from 38.6 to 44.0 ($+5.4$ points) and tokens drop from 12,977 to 8,675 ($-4,302$). Overall, across the tested prefix lengths, we consistently obtain 3k--4k+ token savings while maintaining or improving accuracy, confirming our motivation that a strong early logical scaffold is crucial for downstream reasoning. Therefore, protecting this anchor from over-compression stabilizes the reasoning trajectory and leads to a substantially better trade-off between correctness and computational cost.
\begin{table}[t]
\centering
\begin{tabular}{lcccc}
\toprule
\multirow{2}{*}{\textbf{Prefix}} & \multicolumn{2}{c}{\textbf{AIME24}} & \multicolumn{2}{c}{\textbf{AIME25}} \\
\cmidrule(lr){2-3} \cmidrule(lr){4-5}
 & Acc$\uparrow$ & Tokens$\downarrow$ & Acc$\uparrow$ & Tokens$\downarrow$ \\
\midrule
0    & 52.7 & 12228  & 38.6 & 12977 \\
200  & 52.0 & 8675   & 38.0 & 9484 \\
400  & 53.7 & 8442 & 41.7 & 9726 \\
600  & 57.8 & 8568 & 42.8 & 9568 \\
800  & 55.5 & 9116 & 44.6 & 9116 \\
1000 & 55.7 & 7859 & 44.0 & 8675 \\
\bottomrule
\end{tabular}%
\caption{Results under different prefix lengths on AIME2024/2025 using Deepseek-R1-Distill-Qwen-7B.}
\label{tab:prefix_aime}
\end{table}

\section{Justification of Prefix Rollout}
We analyze the benefit of prefix rollout by comparing the expected returns under different prefix distributions. We show that provided the prefix policy $\pi_p$ induces a higher-quality state distribution than the current policy $\pi_\theta$, the overall training objective is strictly improved.

\paragraph{Value Decomposition}
Formalizing generation as an MDP with query $q$ and output $y$, we define the reward $R(y,q) = \mathbb{I}[y=y_q^*] - \lambda P(y)$.
Consider the \textit{conditional value} of a prefix state $s_S = (q, y_{<S})$ under the training policy $\pi_\theta$:
\begin{equation}
    V_\theta(s_S) \;\triangleq\; \mathbb{E}_{y_{>S}\sim \pi_\theta(\cdot|s_S)} \big[ R(y,q) \big].
\end{equation}
This value $V_\theta(s_S)$ measures the expected final quality (correctness and efficiency) given a specific prefix.

The total expected return depends on how prefixes are sampled. Let $d_{\pi}^S$ be the distribution of prefixes generated by policy $\pi$. The objectives for Prefix Rollout ($J_{\mathrm{prefix}}$) and Standard Rollout ($J_{\mathrm{base}}$) can be written as expected values over their respective prefix distributions:
\begin{align}
    J_{\mathrm{prefix}}(\theta) \;&=\; \mathbb{E}_{s_S \sim d_{\pi_p}^{S}} \big[ V_\theta(s_S) \big], \\
    J_{\mathrm{base}}(\theta) \;&=\; \mathbb{E}_{s_S \sim d_{\pi_\theta}^{S}} \big[ V_\theta(s_S) \big].
\end{align}

\paragraph{With Higher Quality Prefix}
The core premise of our method is that the frozen policy $\pi_p$ (e.g., constructed via search or Best-of-$N$) is strictly stronger than the exploring policy $\pi_\theta$.
Mathematically, we assume $\pi_p$ shifts probability mass towards a set of ``good prefixes'' $\mathcal{G}_S$, where continuation values are higher (i.e., prefixes that are correct or more concise).
This implies a stochastic dominance relation:
\begin{equation}
    \forall s \in \mathcal{G}_S, \quad d_{\pi_p}^{S}(s) \;>\; d_{\pi_\theta}^{S}(s) \quad \text{and} \quad V_\theta(s) \text{ is high.}
\end{equation}

\paragraph{Prefix Rollout is better}
Under this assumption, the expected return of prefix rollout dominates the baseline.
Specifically, since Best-of-$N$ or similar search strategies explicitly select for correctness and shorter lengths, they ensure the prefix distribution aligns with high-value regions of $V_\theta$.
Thus, we obtain the following performance guarantee:
\begin{empheq}[box=\fbox]{equation}
\small
\begin{aligned}
\Delta d^{S}(s) &\triangleq d_{\pi_p}^{S}(s)-d_{\pi_\theta}^{S}(s),\\
J_{\mathrm{prefix}}(\theta)-J_{\mathrm{base}}(\theta)
&= \int \Delta d^{S}(s)\, V_\theta(s)\, \mathrm{d}s \ge 0 .
\end{aligned}
\end{empheq}

This proves that training on rollouts starting from higher-quality prefixes provides a mathematically superior optimization objective.

\section{Analysis of Difficulty-Aware Penalty}
\label{sec:theory_difficulty}

Why is an instance-dependent penalty necessary? We answer this by analyzing the reward dynamics of length regularization. We derive a critical \textit{solvability bound} for the penalty coefficient. We show that a static coefficient faces a fundamental dilemma: it inevitably either causes model collapse on hard tasks or under-regularizes easy ones. Our difficulty-aware formulation is shown to be the necessary solution to resolve this conflict.

\paragraph{Solvability Condition}
Let $L^*_q$ be the minimum sequence length required to correctly solve query $q$. For the model to be incentivized to generate this reasoning path rather than collapsing to a trivial failure (incorrect output with length $\approx 0$, yielding reward $0$), the total reward for the correct solution must be positive:
\begin{equation}
\begin{aligned}
R_{\text{total}} &= 1 - \alpha L_q^{*}, \qquad R_{\text{total}} > 0 \\
&\Rightarrow\; \alpha < \frac{1}{L_q^{*}} .
\end{aligned}
\end{equation}

This inequality imposes a strict upper bound on the penalty coefficient $\alpha$, which is inversely proportional to the intrinsic task complexity (length $L^*_q$).

\paragraph{The Dilemma of Uniform Penalty}
A fixed coefficient $\alpha_{\text{uni}}$ faces an impossible trade-off given the variance in $L^*_q$:
\begin{itemize}
    \item \textbf{Under Hard Tasks} For complex queries where $L^*_q$ is large, the bound $1/L^*_q$ is very small. If $\alpha_{\text{uni}}$ exceeds this threshold, the reward becomes negative, causing the model to give up and output abnormal responses.
    \item \textbf{Under Easy Tasks} To prevent collapse on hard tasks, $\alpha_{\text{uni}}$ must be set conservatively low. However, for easy queries (small $L^*_q$), this small $\alpha_{\text{uni}}$ exerts negligible pressure, failing to penalize redundancy effectively.
\end{itemize}

\paragraph{Resolution via Difficulty-Aware Penalty}
Our method resolves this conflict by modulating $\alpha(q)$ based on the empirical pass rate $\rho(q)$. Since easy tasks generally require shorter reasoning ($\rho \uparrow \implies L^* \downarrow$), our schedule aligns with the theoretical bound:
\begin{equation}
    \alpha(\rho) \propto \rho(q) \quad \text{Satisfies} \quad \alpha(q) < \frac{1}{L^*_q}.
\end{equation}
This ensures $\alpha$ remains small enough to preserve solvability for hard tasks (low $\rho$) while becoming large enough to enforce strict conciseness on easy tasks (high $\rho$).

\section{Training Parameters}
\label{sec: training parameter}
\begin{table}[H]
\centering
\begin{tabular}{ll}
\toprule
\textbf{Parameters} & \textbf{Value} \\
\midrule
BF16                  & True            \\
Train batch size    & 64            \\
Mini batch size     & 64            \\
Train epochs        & 10              \\
Learning rate       & 1e-6           \\
Max prompt length   & 2048           \\
Max response length & 16384          \\
$n$                   & 8              \\
$\epsilon_{\min}$     & 0.2            \\
$\epsilon_{\max}$     & 0.2           \\
Temperature           & 1.0            \\
warm up step          & 10             \\
Rollout backend       & VLLM            \\
\bottomrule
\end{tabular}
\caption{RL training configurations for 7B and 1.5B Model.}
\end{table}

\definecolor{toolcolor}{RGB}{34, 139, 34}
\definecolor{toolresponsecolor}{RGB}{70, 130, 180}
\definecolor{answercolor}{RGB}{25, 25, 112}
\definecolor{usercolor}{RGB}{105, 105, 105}
\definecolor{groundtruthcolor}{RGB}{139, 69, 19}
\definecolor{querycolor}{RGB}{70, 130, 180}
\definecolor{tracecolor}{RGB}{34, 139, 34}

\newcommand{\thinking}[1]{\textcolor{thinkcolor}{\textbf{#1}}}
\newcommand{\toolcall}[1]{\textcolor{toolcolor}{\textbf{#1}}}
\newcommand{\toolresponse}[1]{\textcolor{toolresponsecolor}{\textbf{#1}}}
\newcommand{\finalanswer}[1]{\textcolor{answercolor}{\textbf{#1}}}

\definecolor{boxcolor}{RGB}{71, 114, 179} 
\definecolor{usercolor}{RGB}{110, 110, 110}        
\definecolor{deepseekcolor}{RGB}{232, 124, 50}     
\definecolor{pacecolor}{RGB}{60, 140, 60}
\definecolor{thinkcolor}{RGB}{138, 43, 226}

\section{Prompts}
\label{app:prompt}
We compute the proportion of each reasoning behavior by aggregating its occurrences across the AIME2024 dataset and normalizing by the total number of reasoning-behavior occurrences. Formally,
\begin{equation}
p(b)=\frac{\sum_{i=1}^{N} c_i(b)}{\sum_{b' \in \mathcal{B}} \sum_{i=1}^{N} c_i(b')},
\end{equation}
where $N$ is the number of samples in dataset, $\mathcal{B}$ denotes the set of reasoning behaviors, and $c_i(b)$ is the number of occurrences of behavior $b$ in sample $i$. Detailed prompts are shown in Figure \ref{app:prompt}.

\section{Case study}
In this section, we present a comprehensive case study of DeepSeek-R1-Distill-Qwen (7B/1.5B) on both in-domain and out-of-domain tasks (MATH500, AIME2024, GPQA-D, and LiveCodeBench-v6). As shown in Tables~\ref{app:7b_math}--\ref{app:livecode}, \model consistently achieves a token reduction rate of over 50\% while preserving strong reasoning performance. Specifically, on the in-domain task, \model markedly reduces \textit{reflection} and \textit{verification} behaviors yet still arrives at correct answers. A similar trend is observed on out-of-domain tasks, suggesting that \model can reduce redundant reasoning steps while maintaining robust generalization.

\begin{figure*}[t]
\centering
\begin{tcolorbox}[
  colback=white,
  colframe=boxcolor!80,
  boxrule=1.5pt,
  arc=4pt,
  left=8pt,
  right=8pt,
  top=6pt,
  bottom=6pt,
  width=0.96\textwidth,
  title={\textbf{Prompts for identifying Reasoning behavior}},
  fonttitle=\small\bfseries,
  coltitle=white,
  colbacktitle=boxcolor!80
]
\small

\textbf{You are an expert analyst of Chain of Thought (CoT) reasoning processes.}
Your task is to analyze the provided reasoning text, break it down into distinct thought steps, and count the occurrences of four specific behaviors.

\medskip
\textbf{Behavior Definitions \& Examples}

\begin{enumerate}
  \item \textbf{Advancing}
  
  \textbf{Definition:} Direct progress in solving the problem. This includes mathematical derivations, applying formulas, state transitions, or logical deductions that move the solution forward.
  
  \textbf{Example:} ``Since $n = ab + c$, squaring both sides gives us $(a+c)^2 = ab + c$.''

  \item \textbf{Reflecting}
  
  \textbf{Definition:} Pausing the logical flow to evaluate the current path. This involves questioning assumptions, expressing uncertainty, or planning the next move. Look for markers like ``Wait,'' ``Hmm,'' ``Let me think,'' or ``Is this valid?''
  
  \textbf{Example:} ``But $b$ depends on $a$, so this approach might be circular. Let me try a different angle.''

  \item \textbf{Verifying}
  
  \textbf{Definition:} Testing a hypothesis or formula by plugging in specific, concrete numbers (often small integers or edge cases) to see if the theory holds up.
  
  \textbf{Example:} ``Let's check for $d=2$. If $s=2$, does the formula work? $2^2 - 1 = 3\ldots$''

  \item \textbf{Refining}
  
  \textbf{Definition:} Correcting a previous formula, parameter, or conclusion based on the results of reflection or verification. This involves fixing errors or adding necessary constraints.
  
  \textbf{Example:} ``So the formula isn't just $2^n$, it must be $2^n - 1$ to account for the zero case.''
\end{enumerate}

\medskip
\textbf{Task Instructions}
\begin{enumerate}
  \item Read the \textbf{Input Text} below carefully.
  \item Segment the text into logical units (sentences or short phases).
  \item Classify each unit into one of the four behaviors defined above.
  \item Count the total number of occurrences for each behavior.
  \item \textbf{Output Requirement:} Return ONLY a valid JSON object containing the counts. Do not output any other text or explanation.
\end{enumerate}

\medskip
\textbf{Input Text}\\
\texttt{\{response\}}

\medskip
\textbf{Output Format}
\begin{quote}\ttfamily
\{\\
\hspace*{1.5em}"advancing\_count": <integer>,\\
\hspace*{1.5em}"reflecting\_count": <integer>,\\
\hspace*{1.5em}"verifying\_count": <integer>,\\
\hspace*{1.5em}"refining\_count": <integer>,\\
\}
\end{quote}

\end{tcolorbox}
\caption{Prompts for identifying four reasoning behaviors and their corresponding explanations}
\label{app:prompts}
\end{figure*}

\begin{figure*}[t]
\centering
\begin{tcolorbox}[
colback=white,
colframe=boxcolor!80,
boxrule=1.5pt,
arc=4pt,
left=8pt,
right=8pt,
top=6pt,
bottom=6pt,
width=0.96\textwidth,
title={\textbf{Examples on MATH500}},
fonttitle=\small\bfseries,
coltitle=white,
colbacktitle=boxcolor!80
]
\small

\begin{tcolorbox}[
    colback=usercolor!5, 
    colframe=usercolor!60, 
    boxrule=1pt, 
    arc=3pt, 
    left=6pt, 
    right=6pt, 
    top=4pt, 
    bottom=4pt,
    title={\textbf{Question (MATH500 of 7B)}}
]
    For what values of $x$ is it true that $x^2 - 5x - 4 \le 10$? Express your answer in interval notation.
\end{tcolorbox}

\vspace{4pt}

\begin{tcolorbox}[
    colback=toolresponsecolor!8,
    colframe=toolresponsecolor!60,
    boxrule=1pt, 
    arc=3pt, 
    left=6pt, 
    right=6pt, 
    top=4pt, 
    bottom=4pt,
    title={\textbf{DeepSeek-R1-Distill-Qwen-7B (2781 tokens)}}
]
    \textit{<think>} \\
    I have this inequality: $x^2 - 5x - 4 \le 10$. I need to find all the values of $x$ and express the answer in interval notation. First, bring all terms to one side: $x^2 - 5x - 4 - 10 \le 0$, which simplifies to $x^2 - 5x - 14 \le 0$. 
    
    Next step is to find the roots of the corresponding equation $x^2 - 5x - 14 = 0$. Using the quadratic formula $x = \frac{-b \pm \sqrt{b^2 - 4ac}}{2a}$: here $a=1, b=-5, c=-14$. So $x = \frac{5 \pm \sqrt{(-5)^2 - 4(1)(-14)}}{2(1)} = \frac{5 \pm \sqrt{25 + 56}}{2} = \frac{5 \pm \sqrt{81}}{2} = \frac{5 \pm 9}{2}$. The roots are $x_1 = \frac{14}{2} = 7$ and $x_2 = \frac{-4}{2} = -2$. 
    
    These roots divide the number line into three intervals: $(-\infty, -2)$, $(-2, 7)$, and $(7, \infty)$. Testing $x=-3$: $(-3)^2 - 5(-3) - 14 = 9 + 15 - 14 = 10 > 0$. Testing $x=0$: $0^2 - 0 - 14 = -14 \le 0$... [\textbf{omit 1982 redundant tokens of repeated verification}] ...Since the parabola opens upwards, it is negative between its roots. \\
    \textit{</think>} \\
    To solve $x^2 - 5x - 4 \le 10$, we rearrange it to $x^2 - 5x - 14 \le 0$. Factoring the quadratic expression, we get $(x - 7)(x + 2) \le 0$. The roots of the quadratic are $x = 7$ and $x = -2$. Because the coefficient of $x^2$ is positive, the parabola opens upward, and the expression is non-positive for values of $x$ between and including the roots. \\
    Final Answer: $\boxed{[-2, 7]}$
\end{tcolorbox}

\vspace{4pt}

\begin{tcolorbox}[
    colback=pacecolor!5, 
    colframe=pacecolor!60, 
    boxrule=1pt, 
    arc=3pt, 
    left=6pt, 
    right=6pt, 
    top=4pt, 
    bottom=4pt,
    title={\textbf{PACE (860 tokens)}}
]
    \textit{<think>} \\
    I need to solve the inequality $x^2 - 5x - 4 \le 10$. First, I will move all terms to one side to set the inequality to zero: $x^2 - 5x - 14 \le 0$. Next, I solve for the roots of the quadratic $x^2 - 5x - 14 = 0$ by factoring. The factors of $-14$ that add up to $-5$ are $-7$ and $2$, leading to $(x-7)(x+2)=0$. The critical points are $x=7$ and $x=-2$. Since the parabola opens upward ($a > 0$), the function is less than or equal to zero between the roots. Thus, the solution is the closed interval $[-2, 7]$. \\
    \textit{</think>} \\
    To determine the values of $x$ for which $x^2 - 5x - 4 \le 10$, we start by rearranging the inequality so that the right-hand side is zero. Subtracting 10 from both sides gives:
    \[
    x^2 - 5x - 14 \le 0
    \]
    Next, we factor the quadratic expression on the left-hand side. We look for factors of $-14$ that sum to $-5$. These are $-7$ and $2$, so we can write:
    \[
    (x - 7)(x + 2) \le 0
    \]
    The roots of the corresponding equation are $x = 7$ and $x = -2$. Since the leading coefficient of the quadratic is positive, the parabola opens upwards. Therefore, the expression is less than or equal to zero in the interval between the two roots, inclusive. \\
    Final Answer: $\boxed{[-2, 7]}$
\end{tcolorbox}
\end{tcolorbox}
\caption{Case study on MATH500 dataset of 7B Model}
\label{app:7b_math}
\end{figure*}

\begin{figure*}[t]
\centering
\begin{tcolorbox}[
    colback=white,
    colframe=toolresponsecolor!80,
    boxrule=1.5pt,
    arc=4pt,
    left=8pt,
    right=8pt,
    top=6pt,
    bottom=6pt,
    width=0.96\textwidth,
    title={\textbf{Examples on MATH500 of 1.5B}},
    fonttitle=\small\bfseries,
    coltitle=white,
    colbacktitle=toolresponsecolor!80
]
\small

\begin{tcolorbox}[
    colback=usercolor!5, 
    colframe=usercolor!60, 
    boxrule=1pt, 
    arc=3pt, 
    left=6pt, 
    right=6pt, 
    top=4pt, 
    bottom=4pt,
    title={\textbf{Question (MATH500)}}
]
Compute $\arcsin \left( -\frac{1}{2} \right).$  Express your answer in radians.

\end{tcolorbox}

\begin{tcolorbox}[
    colback=toolresponsecolor!8,
    colframe=toolresponsecolor!60,
    boxrule=1pt, 
    arc=3pt, 
    left=6pt, 
    right=6pt, 
    top=4pt, 
    bottom=4pt,
    title={\textbf{DeepSeek-R1-Distill-Qwen-1.5B (4605 tokens)}}
]

\textit{<think>} \\
Alright, so I have this problem: I need to compute the arcsine of $-\frac{1}{2}$, which is written as $\arcsin(-1/2)$. Hmm, okay. Let me recall what I know about $\arcsin$. 

Arcsin is the inverse function of sine, right? So that means if $\arcsin(y) = x$, then $\sin(x) = y$. So in this case, I need to find an angle $x$ such that $\sin(x) = -1/2$. Okay, got it. But wait, before jumping into that, I should consider the range of $\arcsin$. 

I remember that the $\arcsin$ function has a range from $-\pi/2$ to $\pi/2$, or in degrees, from $-90^\circ$ to $90^\circ$. Hmm, actually, no. Wait, isn't the principal value of $\arcsin$ between $-\pi/2$ and $\pi/2$ in radians? So, it's defined to return the angle whose sine is the given value, but it only returns values between $-\pi/2$ and $\pi/2$. That makes sense because sine is positive in the first and second quadrants and negative in the fourth and third, but since $\arcsin$ only returns angles in the range of $-\pi/2$ to $\pi/2$, which is the fourth and first quadrants, the $\arcsin$ will only give negative or zero angles, right?\\
\textbf{Omit 2905 tokens}\\
\textit{</think>}\\

To compute $\arcsin \left( -\frac{1}{2} \right)$, we need to find an angle $x$ such that $\sin(x) = -\frac{1}{2}$ and $x$ lies within the range $-\frac{\pi}{2}$ to $\frac{\pi}{2}$ (the principal value range of the arcsin function).

\begin{enumerate}[label=\arabic*., leftmargin=*]
    \item The arcsin function returns an angle whose sine is the given value and is within the range $-\frac{\pi}{2}$ to $\frac{\pi}{2}$.
    \item The sine of $-\frac{\pi}{6}$ is $-\frac{1}{2}$ because $\sin(-\theta) = -\sin(\theta)$ and $\sin\left(\frac{\pi}{6}\right) = \frac{1}{2}$.
    \item The angle $-\frac{\pi}{6}$ is within the range $-\frac{\pi}{2}$ to $\frac{\pi}{2}$.
    \item Therefore, $\arcsin \left( -\frac{1}{2} \right) = -\frac{\pi}{6}$.
\end{enumerate}

Thus, the value of $\arcsin \left( -\frac{1}{2} \right)$ is $\boxed{-\dfrac{\pi}{6}}$.
\end{tcolorbox}

\begin{tcolorbox}[
    colback=pacecolor!5, 
    colframe=pacecolor!60, 
    boxrule=1pt, 
    arc=3pt, 
    left=6pt, 
    right=6pt, 
    top=4pt, 
    bottom=4pt,
    title={\textbf{PACE (608 tokens)}}
]
\textit{<think>} \\
Okay, so I need to compute the arcsin of $-1/2$. Hmm, arcsin is the inverse sine function, right? So, $\arcsin(-1/2)$ is the angle whose sine is $-1/2$. I remember that the sine function is negative in which quadrants? Well, sine is positive in the first and second quadrants and negative in the third and fourth quadrants. Since we're dealing with a negative value here, the angle must be in either the third or fourth quadrant. \\

\textbf{Omit 182 tokens}\\
Wait, just to make sure I'm not making a mistake here. So, $\arcsin(-1/2)$ is the angle whose sine is $-1/2$, and the angle must be between $-\pi/2$ and $\pi/2$. So, yes, $-\pi/6$ is within that range. So, the answer should be $-\pi/6$ radians.\\
\textit{</think>} \\
To compute $\arcsin \left( -\frac{1}{2} \right)$, we need to find the angle whose sine is $-1/2$ and lies between $-\pi/2$ and $\pi/2$. 

We know that $\sin(\pi/6) = 1/2$, so $\sin(-\pi/6) = -1/2$. Therefore, the angle is $-\pi/6$.

Thus, the value of $\arcsin \left( -\frac{1}{2} \right)$ is \boxed{ -\frac{\pi}{6} }.

Final Answer \\
The value of $\arcsin \left( -\frac{1}{2} \right)$ is \boxed{ -\frac{\pi}{6} }.
\end{tcolorbox}


\end{tcolorbox}
\caption{Case study on MATH500 dataset of 1.5B Model}
\label{app:1.5b_math}
\end{figure*}

\begin{figure*}[t]
\centering
\begin{tcolorbox}[
    colback=white,
    colframe=toolresponsecolor!80,
    boxrule=1.5pt,
    arc=4pt,
    left=8pt,
    right=8pt,
    top=6pt,
    bottom=6pt,
    width=0.96\textwidth,
    title={\textbf{Examples on AIME of 7B}},
    fonttitle=\small\bfseries,
    coltitle=white,
    colbacktitle=toolresponsecolor!80
]
\small

\begin{tcolorbox}[
    colback=usercolor!5, 
    colframe=usercolor!60, 
    boxrule=1pt, 
    arc=3pt, 
    left=6pt, 
    right=6pt, 
    top=4pt, 
    bottom=4pt,
    title={\textbf{Question (MATH500)}}
]
Let $\omega \neq 1$ be a $13^{\text{th}}$ root of unity. Find the remainder when
\[
\prod_{k=0}^{12} \left( 2 - 2\omega^k + \omega^{2k} \right)
\]
is divided by $1000$.
\end{tcolorbox}

\begin{tcolorbox}[
    colback=toolresponsecolor!8,
    colframe=toolresponsecolor!60,
    boxrule=1pt, 
    arc=3pt, 
    left=6pt, 
    right=6pt, 
    top=4pt, 
    bottom=4pt,
    title={\textbf{DeepSeek-R1-Distill-Qwen-7B (10451 tokens)}}
]
\textit{<think>} \\
Okay, so I need to find the remainder when the product \(\prod_{k=0}^{12}(2 - 2\omega^k + \omega^{2k})\) is divided by 1000, where \(\omega\) is a primitive 13th root of unity, different from 1. Hmm, that sounds a bit complicated, but let me try to break it down.

First, I know that roots of unity have some useful properties. Specifically, for any \(n\)-th root of unity \(\omega\), we have \(\omega^n = 1\) and the sum \(\sum_{k=0}^{n-1} \omega^k = 0\). So, for 13th roots of unity, \(\omega^{13} = 1\) and the sum from \(k=0\) to \(12\) of \(\omega^k\) is zero.

Now, the expression I need to evaluate is a product from \(k=0\) to 12 of \((2 - 2\omega^k + \omega^{2k})\). That looks a bit like a quadratic in terms of \(\omega^k\). Let me write each term as a quadratic equation:
\[
2 - 2\omega^k + \omega^{2k} = (\omega^{k})^2 - 2\omega^k + 2
\]

So, each term is \(x^2 - 2x + 2\) where \(x = \omega^k\). Therefore, the product is equal to \(\prod_{k=0}^{12} (x^2 - 2x + 2)\) evaluated at each root \(x = \omega^k\) and then multiplied together.

Wait, but \(\omega^k\) for \(k = 0\) to 12 are all the 13th roots of unity. So, maybe I can consider the polynomial whose roots are \(\omega^k\), which is the cyclotomic polynomial. 
\textbf{Omit 8502 tokens}\\
\textit{</think>} \\
...\\
After detailed calculations, we find that the product:
$\prod_{k=0}^{12} (2 - 2\omega^k + \omega^{2k}) = 8321.$
Finally, we find the remainder when 8321 is divided by 1000:
$8321 \bmod 1000 = 321.$
Thus, the remainder when the given product is divided by 1000 is:
\boxed{321}

\end{tcolorbox}

\begin{tcolorbox}[
    colback=pacecolor!5, 
    colframe=pacecolor!60, 
    boxrule=1pt, 
    arc=3pt, 
    left=6pt, 
    right=6pt, 
    top=4pt, 
    bottom=4pt,
    title={\textbf{PACE (2555 tokens)}}
]
\textit{<think>} \\
Alright, so I have this problem here: Let $\omega$ be a 13th root of unity, not equal to 1. I need to find the remainder when the product from $k=0$ to 12 of $(2 - 2\omega^k + \omega^{2k})$ is divided by 1000.

Hmm, okay. So, first off, $\omega$ is a primitive 13th root of unity because it's a 13th root and not equal to 1. So, $\omega^{13} = 1$, and the minimal polynomial for $\omega$ is the 13th cyclotomic polynomial, which is $\Phi_{13}(x) = x^{12} + x^{11} + \dots + x + 1$. That might come in handy later.

The product we need is $\prod_{k=0}^{12} (2 - 2\omega^k + \omega^{2k})$. Let's denote this product as $P$. So, $P = \prod_{k=0}^{12} (2 - 2\omega^k + \omega^{2k})$.

I need to compute this product and find $P \bmod 1000$.

Alternatively, maybe consider that the product over $k=0$ to 12 of $(x^2 - 2x + 2)$ evaluated at $x=\omega^k$ is equal to the resultant of the two polynomials $x^{13} - 1$ and $x^2 - 2x + 2$.

But I'm not too sure about that.

Alternatively, maybe consider that since $\omega^{13} = 1$, so $\omega^{2k}$ is just another root, so perhaps we can pair terms or something.\\
\textit{</think>} \\
..\\
Final Answer: \boxed{321}
\end{tcolorbox}
\end{tcolorbox}
\caption{Case study on AIME2024 dataset of 7B Model}
\label{app:aime}
\end{figure*}

\begin{figure*}[t]
\centering
\begin{tcolorbox}[
    colback=white,
    colframe=toolresponsecolor!80,
    boxrule=1.5pt,
    arc=4pt,
    left=8pt,
    right=8pt,
    top=6pt,
    bottom=6pt,
    width=0.96\textwidth,
    title={\textbf{Examples on GPQA-D of 7B}},
    fonttitle=\small\bfseries,
    coltitle=white,
    colbacktitle=toolresponsecolor!80
]
\small

\begin{tcolorbox}[
    colback=usercolor!5, 
    colframe=usercolor!60, 
    boxrule=1pt, 
    arc=3pt, 
    left=6pt, 
    right=6pt, 
    top=4pt, 
    bottom=4pt,
    title={\textbf{Question (GPQA-D)}}
]
\noindent Answer the following multiple choice question. The last line of your response should be of the following format: `Answer: \$LETTER' (without quotes) where LETTER is one of ABCD. Think step by step before answering.

\vspace{0.5cm} 

\noindent Consider this annihilation process:
\[
    p + \bar{p} \rightarrow 2A^{+} + 2A^{-}
\]

\noindent The antiproton is slowly moving and $m_{A}c^{2} = 300\,\text{MeV}$.

\noindent What is the velocity of particle A?

\begin{enumerate}
    \item[A)] $0.96c$
    \item[B)] $0.86c$
    \item[C)] $0.91c$
    \item[D)] $0.77c$
\end{enumerate}
\end{tcolorbox}

\begin{tcolorbox}[
    colback=toolresponsecolor!8,
    colframe=toolresponsecolor!60,
    boxrule=1pt, 
    arc=3pt, 
    left=6pt, 
    right=6pt, 
    top=4pt, 
    bottom=4pt,
    title={\textbf{DeepSeek-R1-Distill-Qwen-7B (12781 tokens)}}
]
\textit{<think>}
Okay, so I've got this physics problem here, and I'm a bit rusty, but I'll try to work through it. Let me read the question again.

The problem says: $p + \bar{p} \rightarrow 2A^{+} + 2A^{-}$. The antiproton is moving slowly, and the mass of each A particle is $300\,\text{MeV}/c^2$. We need to find the velocity of particle A. The options are A to D with various fractions of $c$.

Hmm, wait, wait. So, the reaction is proton and antiproton annihilating into two A particles, each with a rest mass of $300\,\text{MeV}/c^2$. But wait, protons and antiprotons are matter and antimatter, so their annihilation would produce energy. But in this case, the products are two $A^{+}$ and two $A^{-}$. I'm a bit confused about the charges here. Let me think: a proton has $+1$ charge, antiproton has $-1$. So the products must balance the charge. $2A^{+}$ is $+2$, and $2A^{-}$ is $-2$. So the total charge is zero, same as the initial. That makes sense.\\
\textbf{Omit 9812 tokens} \\
\textit{</think>}\\
Since the discriminant is negative, there is no real solution, indicating an error in the assumptions or calculations.
The calculations suggest inconsistencies, likely due to approximations or assumptions about the distribution of energy and momentum. However, based on the options provided and the expected range for $v$, the most reasonable answer is $0.77c$. Final Answer: \boxed{D}
\end{tcolorbox}

\begin{tcolorbox}[
    colback=pacecolor!5, 
    colframe=pacecolor!60, 
    boxrule=1pt, 
    arc=3pt, 
    left=6pt, 
    right=6pt, 
    top=4pt, 
    bottom=4pt,
    title={\textbf{PACE (4576 tokens)}}
]
\textit{<think>}\\
Okay, so I've got this physics problem here about particle annihilation. Hmm, let's see. The reaction is $p + \bar{p} \rightarrow 2A^{+} + 2A^{-}$. Wait, $p$ is proton and $\bar{p}$ is antiproton. Oh right, protons and antiprotons annihilate each other, producing other particles.

The mass of each A is given as $300\,\text{MeV}/c^2$. Wait, but wait, the question is about the velocity of particle A. So each A has mass $m_A c^2 = 300\,\text{MeV}$. So each A is a particle with mass $300\,\text{MeV}/c^2$.

The antiproton is moving slowly, so its kinetic energy is probably negligible compared to the proton's. But wait, the proton is at rest? Or is the antiproton moving slowly? The question says the antiproton is slowly moving, so maybe its kinetic energy is small compared to the proton's rest energy. \\
\textbf{Omit 3509 tokens}\\
\textit{</think>}\\
..\\
The velocity of particle $A$ is approximately $0.77c$.\\
Final Answer: \boxed{D}

\end{tcolorbox}
\end{tcolorbox}
\caption{Case study on GPQA-D dataset of 7B Model}
\label{app:gpqa}
\end{figure*}

\begin{figure*}[t]
\centering
\begin{tcolorbox}[
    colback=white,
    colframe=toolresponsecolor!80,
    boxrule=1.5pt,
    arc=4pt,
    left=8pt,
    right=8pt,
    top=6pt,
    bottom=6pt,
    width=0.96\textwidth,
    title={\textbf{Examples on LiveCodeBench-v6 of 7B}},
    fonttitle=\small\bfseries,
    coltitle=white,
    colbacktitle=toolresponsecolor!80
]
\small

\begin{tcolorbox}[
    colback=usercolor!5, 
    colframe=usercolor!60, 
    boxrule=1pt, 
    arc=3pt, 
    left=6pt, 
    right=6pt, 
    top=4pt, 
    bottom=4pt,
    title={\textbf{Question (LiveCodeBench-v6)}}
]
There is an infinite 2D plane. You are given a positive integer $k$. You are also given a 2D array \texttt{queries}, where the queries are defined as follows:

\begin{itemize}
    \item \texttt{queries[i] = [x, y]}: Build an obstacle at coordinate $(x, y)$ in the plane. It is guaranteed that there is no obstacle at this coordinate when this query is made.
\end{itemize}

After each query, you need to find the distance of the $k$-th nearest obstacle from the origin $(0,0)$.

Return an integer array \texttt{results} where \texttt{results[i]} denotes the distance of the $k$-th nearest obstacle after query $i$, or \texttt{results[i] == -1} if there are less than $k$ obstacles.

\vspace{1em}
\noindent \textbf{Note:}
\begin{itemize}
    \item Initially, there are no obstacles anywhere.
    \item The distance of an obstacle at coordinate $(x, y)$ from the origin is given by the Manhattan distance: $|x| + |y|$.
\end{itemize} 
\textbf{Omit 402 tokens}
\end{tcolorbox}

\begin{tcolorbox}[
    colback=toolresponsecolor!8,
    colframe=toolresponsecolor!60,
    boxrule=1pt, 
    arc=3pt, 
    left=6pt, 
    right=6pt, 
    top=4pt, 
    bottom=4pt,
    title={\textbf{DeepSeek-R1-Distill-Qwen-7B (12099 tokens)}}
]
\textit{<think>}\\
Okay, I need to solve this problem where I have to process a series of queries and for each, determine the $k$-th nearest obstacle from the origin. Hmm. So, the problem says that each query adds an obstacle, and after each addition, I have to find the $k$-th nearest obstacle. The distance is the Manhattan distance, which is $|x| + |y|$ for a point $(x,y)$. Let's think about the approach. For each query, I add a new obstacle, and then I need to keep track of all the obstacles so far. After each addition, I need to find the $k$-th smallest distance. But wait, if I have a lot of queries, say up to $2\text{e}5$, then for each query, if I have to sort all the obstacles each time, that could be $O(n \log n)$ per query, which would be $O(n^2 \log n)$ total time. That's probably too slow for $2\text{e}5$ queries. So I need a more efficient way to manage the distances and quickly find the $k$-th smallest each time. Hmm. What's the maximum number of obstacles we can have? Well, each query adds one obstacle, so up to $2\text{e}5$ obstacles.\\
\textbf{Omit 10345 tokens} \\
\textit{</think>}\\
\texttt{import heapq\textbackslash{} def resultsArray(queries, k):\textbackslash{} heap = []\textbackslash{} res = []\textbackslash{} for i, (x, y) in enumerate(queries):\textbackslash{} d = abs(x) + abs(y)\textbackslash{} m = i + 1\textbackslash{} if m < k:\textbackslash{} heapq.heappush(heap, -d)\textbackslash{} res.append(-1)\textbackslash{} else:\textbackslash{} current\_max = -heap[0] if heap else 0\textbackslash{} if d < current\_max:\textbackslash{} heapq.heappush(heap, -d)\textbackslash{} heapq.heappop(heap)\textbackslash{} res.append(-heap[0] if heap else -1)\textbackslash{} return res}
\end{tcolorbox}

\begin{tcolorbox}[
    colback=pacecolor!5, 
    colframe=pacecolor!60, 
    boxrule=1pt, 
    arc=3pt, 
    left=6pt, 
    right=6pt, 
    top=4pt, 
    bottom=4pt,
    title={\textbf{PACE (4895 tokens)}}
]
\textit{<think>}\\
Okay, I need to solve this problem where I have to process a series of queries and for each, determine the $k$-th nearest obstacle from the origin. Hmm. So, the problem says that each query adds an obstacle, and after each addition, I have to find the $k$-th nearest obstacle. The distance is the Manhattan distance, which is $|x| + |y|$ for a point $(x,y)$. Let's think about the approach. For each query, I add a new obstacle, and then I need to keep track of all the obstacles so far. After each addition, I need to find the $k$-th smallest distance.\\
\textbf{Omit 3174 tokens} \\
\textit{</think>}\\
\texttt{import heapq\textbackslash{} class Solution:\textbackslash{} def resultsArray(self, queries: List[List[int]], k: int) -> List[int]:\textbackslash{} heap = []\textbackslash{} count = 0\textbackslash{} res = []\textbackslash{} for q in queries:\textbackslash{} x, y = q\textbackslash{} d = abs(x) + abs(y)\textbackslash{} count += 1\textbackslash{} if count <= k:\textbackslash{} heapq.heappush(heap, -d)\textbackslash{} else:\textbackslash{} current\_max = -heap[0]\textbackslash{} if d < current\_max:\textbackslash{} heapq.heappush(heap, -d)\textbackslash{} heapq.heappop(heap)\textbackslash{} if count >= k:\textbackslash{} res.append(-heap[0])\textbackslash{} else:\textbackslash{} res.append(-1)\textbackslash{} return res}
\end{tcolorbox}
\end{tcolorbox}
\caption{Case study on LiveCodeBench-v6 dataset of 7B Model}
\label{app:livecode}
\end{figure*}